\pdfoutput=1

\documentclass[11pt]{article}

\usepackage{acl}

\usepackage{times}
\usepackage{latexsym}
\usepackage{microtype}
\usepackage{inconsolata}
\usepackage{graphicx}
\usepackage{booktabs}       
\usepackage{amsmath}
\usepackage{amssymb}
\usepackage{bm}             
\usepackage{multirow}
\usepackage{pifont}        
\usepackage{xcolor}
\usepackage{url}
\usepackage{hyperref}
\usepackage{tikz}           
\usetikzlibrary{arrows.meta, positioning, shapes.geometric, fit, backgrounds}

\title{Morpheus: A Morphology-Aware Neural Tokenizer and Word Embedder for Turkish}

\author{
  Şakar, Tolga \\
  Independent Researcher \\
  \texttt{lonewolf\_rd@protonmail.com}
}

\begin{document}
\maketitle

\begin{abstract}
Turkish is agglutinative: meaning is carried by morphemes, yet the subword
tokenizers that drive modern language models segment words by corpus statistics,
and several of them---through Unicode and case normalization, whitespace
handling, or allomorph canonicalization---cannot decode their output back to the
input text. This paper presents \textbf{Morpheus}, a neural morpheme-boundary
model for Turkish that is at once a reversible, morphology-aware tokenizer and a
word embedder. A differentiable Poisson--binomial dynamic program turns
per-character boundary probabilities into soft morpheme memberships during
training and exact segments at inference; together with a lossless
pre-tokenizer and byte fallback, $\mathrm{decode}(\mathrm{encode}(x)) = x$ holds
for any input text. Morpheus is compared with case-preserving BPE, Unigram,
WordPiece, byte-level BPE, Morfessor, and a rule-based Turkish tokenizer.
Morpheus and byte-level BPE are the only tokenizers that reconstruct every input
exactly, from $30{,}204$ inflected words to $5{,}000$ raw corpus lines and a stress
set of code, emoji, and irregular whitespace. Among these reversible tokenizers
only Morpheus produces morpheme-aligned units: MorphScore macro-F1 $0.61$ versus
$0.35$--$0.44$ for the subword family, a TR-MMLU pure-token rate of $57\%$ versus
$32$--$33\%$, and exact segmentation of $70\%$ of rare roots versus at most $45\%$.
These gains have a measured cost: Morpheus emits $30\%$ more tokens than
byte-level BPE on the same text, and in a controlled language-modeling probe
(a $58$M-parameter model trained with equal compute) its token stream reaches
$1.451$ bits per character against $1.411$. As an embedder, frozen
$320$-dimensional Morpheus vectors lead root-family retrieval over $120$ gold
families (MAP $0.89$ versus $0.72$ for BGE-M3 and $0.38$ for BERTurk) and
same-root verification (ROC-AUC $0.999$), while carrying little number and almost
no case information---the expected result of a root-centric training objective.
Code:
\url{https://github.com/lonewolf-rd/TurkishMorpheus}; model:
\url{https://huggingface.co/lonewolflab/Morpheus-TR-50K}; interactive demo:
\url{https://huggingface.co/spaces/lonewolflab/morpheus-tr-demo}.
\end{abstract}

\section{Introduction}

Turkish is an agglutinative language that encodes most of its semantic content
in productive chains of derivational and inflectional suffixes attached to a
root; a single root can unfold into hundreds of distinct surface forms through
the ordering of its morphemes (e.g.\ \textit{ev} ``house'' $\rightarrow$
\textit{evlerimizdekiler} ``the ones in our houses''). The unit that carries
meaning in Turkish is therefore the morpheme, not the word and not a
frequency-driven fragment of it. This property places two distinct demands on the
machinery of modern Turkish NLP---one on \emph{tokenization} and one on
\emph{word representation}---and, as argued below, current tools meet each of
them only partially.

\paragraph{The tokenization problem.}
Subword tokenizers such as BPE, WordPiece, and Unigram
\cite{sennrich2016bpe,kudo2018sentencepiece} segment words by corpus statistics
rather than morphology, and on Turkish this produces two concrete failures.
First, many tokenizers are \emph{not reversible}: decoding the ids does not
recover the original string. Whether a tokenizer is reversible depends less on
its segmentation algorithm than on its preprocessing---Unicode normalization,
case folding, accent stripping, whitespace collapsing, and the canonicalization
of allomorphs each discard surface information. In a generative language model,
where every generated id must decode to faithful text, such losses corrupt the
output and silently degrade any downstream use of the decoded string. Byte-level
BPE \citep{radford2019gpt2}, which operates on raw UTF-8 bytes, avoids the
problem entirely. Second, all frequency-driven methods, byte-level BPE included,
cut words at positions unrelated to their morphology: frequent words are kept
whole, and rarer ones are split into fragments that do not correspond to the units
that carry meaning. Unsupervised morphological segmenters such as Morfessor
\cite{creutz2007morfessor} and rule-based analyzers such as Zemberek
\cite{akin2007zemberek} address the morphological side, but the former is not
optimized for language modeling and the latter is lossy and dictionary-bound. In
short, existing tokenizers offer either exact reversibility or morphological
alignment, but not both; and morphological alignment comes at the price of more
tokens per word.

\paragraph{The representation problem.}
The same morphological richness also strains Turkish word representation.
Contextual encoders such as BERTurk \cite{schweter2020berturk} provide strong
embeddings, but they are heavyweight ($\sim$110M+ parameters), tied to their own
subword vocabularies, and treat morphology only implicitly. A representation in
which morphologically related forms (\textit{kitap}, \textit{kitaplar},
\textit{kitab\i m\i z}) sit together by construction---rather than only after
large-scale pretraining---remains absent. More fundamentally, tokenization and
representation are currently solved by \emph{two separate systems}: a tokenizer
produces discrete ids that carry no meaning, and a distinct, much larger model
must be trained to turn those ids into vectors. For an agglutinative language,
where the boundary information needed to tokenize well and the structure needed
to represent well are one and the same morphological signal, this separation is
wasteful.

\paragraph{This paper.}
These gaps motivate a single Turkish model that is simultaneously a
\emph{reversible, morphology-aware tokenizer} and a \emph{structured
word-embedding producer}. This paper introduces \textbf{Morpheus}, a neural
morpheme-boundary model for Turkish that provides both. Morpheus combines
boundary supervision from an unsupervised analyzer (Morfessor) with
self-supervised objectives (skip-gram negative sampling, root-family contrastive
learning, and masked language modeling), and segments words through a
differentiable Poisson--binomial dynamic program: gradients flow over soft
morpheme memberships during training, while inference recovers exact hard
boundaries with no architectural switch and no string normalization. Because no
normalization is applied, the emitted pieces \emph{are} the surface form; a
lossless pre-tokenizer and a byte-fallback id mapping extend this from words to
arbitrary text, so $\mathrm{decode}(\mathrm{encode}(x)) = x$ holds by
construction. Because the model is neural, the same forward pass that tokenizes
also yields a structured $\mathbb{R}^{320}$ embedding per word. This paper
evaluates Morpheus in these two roles---as a tokenizer and as a word
embedder---against established alternatives, and reports what it gains and what
it costs in each. Integrating Morpheus into a full Turkish language model is the
subject of follow-up work.

The contributions of this paper are:
\begin{itemize}
    \item \textbf{Morpheus}, a neural morphology-aware tokenizer for Turkish that
    is exactly reversible on raw text without inference-time normalization, via a
    differentiable Poisson--binomial segmentation that unifies training and
    inference and a byte-fallback id mapping that covers any input character.
    \item A demonstration that the \emph{same model} is a word-embedding
    producer, evaluated against a contextual encoder (BERTurk) and a multilingual
    retriever (BGE-M3) on root-family retrieval, same-root verification,
    morphological probing, and Turkish NER.
    \item An evaluation in which every baseline preserves case, covering
    reversibility at three levels (isolated words, raw text, a stress set), gold
    morphological alignment (MorphScore, SIGMORPHON), surface fidelity, token
    validity (TR-MMLU), language-modeling BPC with a held-out validation split,
    and inference efficiency.
    \item A characterization of the trade-off between morphological alignment
    and sequence length, comparing Morpheus with byte-level BPE, the other
    exactly reversible tokenizer, property by property.
\end{itemize}

\paragraph{Revision note.}
This version corrects and extends version~1. The tokenizer now preserves all
text (punctuation, case, whitespace) and falls back to bytes for unseen
characters; every baseline was retrained without case folding, and the
byte-level baseline is now a true byte-level BPE. All experiments were re-run on
a rebuilt corpus. Two claims of version~1 do not hold under these conditions:
Morpheus is not the only exactly reversible tokenizer (byte-level BPE is
reversible too), and it does not reach the lowest bits per character among
reversible tokenizers. Version~1's BPC result was obtained with a pre-tokenizer
that removed punctuation from Morpheus's token stream (Section~\ref{sec:lm}).
The root-family retrieval evaluation now uses $120$ gold families instead of a
small hand-picked set.

\section{Related Work}

\paragraph{Subword tokenization and its limits for Turkish.}
BPE \citep{sennrich2016bpe}, WordPiece \citep{devlin2019bert}, and Unigram
\citep{kudo2018subword}, implemented at scale through SentencePiece
\citep{kudo2018sentencepiece}, are the de facto interface between text and
modern language models; GPT-2 introduced byte-level BPE
\citep{radford2019gpt2}, which merges raw UTF-8 bytes and is therefore reversible
on any input. A growing body of work shows that this frequency-driven
design is not neutral for morphologically rich languages such as Turkish.
\citet{toraman2023impact} compare five tokenizers at different granularities and
find that a morphological-level tokenizer is competitive with the de facto
ones while responding more strongly to vocabulary size, and that the ratio of
vocabulary to model parameters is itself a design variable. \citet{kaya2024granularity}
study vocabulary size for Turkish BERT models across NER, sentiment, and QA, and
\citet{altinok2026optimal} present a systematic evaluation of the
data--vocabulary--morphology interplay under matched parameter budgets, together
with morphology-aware diagnostics (boundary F1, lemma atomicity,
over-/under-segmentation). These studies quantify the cost of frequency-driven
segmentation; Morpheus instead attacks it at the source, by learning morpheme
boundaries with a neural model.

\paragraph{Morphology-aware and linguistically informed tokenizers.}
The unsupervised Morfessor family \citep{creutz2002unsupervised,creutz2007morfessor}
induces morpheme-like units via a minimum-description-length objective and
remains a standard segmentation baseline for agglutinative languages; here it serves
as the boundary teacher for Morpheus. Rule-based analyzers such as Zemberek
\citep{akin2007zemberek} encode Turkish morphology explicitly but are
dictionary-bound. More recent Turkish-specific tokenizers improve linguistic
alignment in different ways: \citet{bayram2025twm} propose a hybrid tokenizer
(TurkishTokenizer) that combines dictionary-driven root/affix segmentation,
phonological normalization mapping allomorphic variants to shared identifiers,
and a subword fallback, reporting strong Turkish-token and purity rates and
competitive STS and TurBLiMP results; \citet{gulgonul2025hece} exploit the
closed syllable inventory of Turkish for a resource-light, retrieval-oriented
tokenizer. These methods raise morphological alignment, but they do so through
runtime normalization (which discards surface information, e.g.\ mapping
allomorphs to a canonical id) or through fixed dictionaries and syllable
inventories. Morpheus differs on two axes: it learns boundaries neurally rather
than from a lexicon, and it applies no normalization, so segmentation is
surface-preserving and exactly invertible---while, uniquely, the same model also
yields word embeddings.

\paragraph{Evaluation standards for Turkish tokenization.}
\citet{bayram2025standards} and its conference counterpart \citep{bayram2025siu}
introduce the TR-MMLU benchmark and the Turkish-token (\%TR) and pure-token
(\%Pure) metrics, arguing that linguistic alignment of tokens correlates with
downstream performance more strongly than raw token purity. This work adopts the
\%TR/\%Pure protocol for vocabulary-level comparison and complement it with
metrics that prior comparisons largely omit: exact reversibility, gold
morpheme F1 (MorphScore), SIGMORPHON inflection alignment, surface-string
fidelity, and bits-per-character under a parameter-equalized language-model
budget. Together these make explicit the lossless-versus-lossy axis that
separates tokenizers valid for generation from those that are not, and, within
the lossless group, how far the units align with morphology.

\paragraph{Turkish word representations and the tokenizer--embedding gap.}
On the representation side, BERTurk \citep{schweter2020berturk} provides strong
contextual Turkish embeddings, and recent work adapts multilingual encoders to
Turkish---e.g.\ \citet{bayram2026embedding} perform cross-lingual tokenizer
surgery and offline distillation to build a Turkish sentence-embedding model,
while general multilingual retrievers such as BGE-M3 \citep{chen2024bgem3} are
competitive on Turkish out of the box. All of these treat representation as a
system separate from---and much larger than---the tokenizer. Morpheus instead
couples the two: a single neural model both tokenizes losslessly and emits a
morphology-derived embedding, and that embedding is evaluated directly against
BERTurk and BGE-M3.

\section{Methodology}

\subsection{Data and preprocessing}
Morpheus is trained on a large-scale monolingual Turkish corpus that combines a
multi-register author corpus with the full cleaned Turkish Wikipedia
($\sim$10\,GB of raw text), assembled to expose the model to diverse
morphological constructions across four registers: Ekşisözlük
(informal/colloquial, rich in spoken-language suffixation), Dergipark (academic,
derivational morphology and terminology), Turkish news sites (standard
journalistic), and Turkish Wikipedia (encyclopedic, broad vocabulary). The
web-sourced registers were collected and cleaned with a companion scraping
toolkit that documents per-source extraction, HTML/URL stripping, Unicode
normalization, and deduplication; the Wikipedia portion is additionally filtered
for Turkish-alphabet coverage, stopword/length thresholds, and markup, then
deduplicated. All text is processed with Turkish-aware case folding
($\textit{\.I}\!\rightarrow\!\textit{i}$, $\textit{I}\!\rightarrow\!\textit{\i}$),
with the original casing retained as a per-character side channel rather than
discarded.

\subsection{Caching, supervision, and splits}
The corpus is split $95/5$ into train and test partitions with a fixed seed. To
remove per-epoch segmentation overhead, each sentence is pre-tokenized once into
a cached tensor bundle containing, per word: character ids (padded to
$\text{max\_word\_len}=32$), per-character case flags, a
$(\text{max\_word\_len}-1)$ binary boundary-label vector from the Morfessor
teacher, a word id against a $120$K word vocabulary, and a root id against a
$30$K root vocabulary (the root being the first Morfessor segment), together
with a sentence attention mask. The boundary labels are produced by Morfessor
\citep{creutz2007morfessor} and then \emph{root-corrected}: for in-dictionary
words, intra-root Morfessor boundaries are removed when an independent root
lexicon agrees on the root span, reducing root over-segmentation. This
correction is applied only to the training labels and is purely positional---it
never rewrites strings---so Morpheus remains surface-preserving at inference.
For Morpheus training the sentence cache is capped at $900$K (train) / $100$K
(validation) sentences, while the word and root vocabularies are built from the
full corpus; the separate $1$M-line cap referred to later applies only to the
downstream language-model evaluation (Section~\ref{sec:lm}), not to Morpheus
itself. The Morpheus checkpoint and its Morfessor teacher were trained on an
earlier snapshot of the corpus; all other components were built on a rebuilt
snapshot (Section~\ref{sec:limits}).

\subsection{Model architecture}
Morpheus maps a word, given as a character sequence, to a set of morpheme
boundaries and a single word embedding in one forward pass, through three stages
connected by a differentiable segmentation operator. All hidden states share a
working dimension of $d=320$.

\paragraph{Character encoder and positional morphology.}
Each character embedding is concatenated with a learned case-flag embedding,
passed through a multi-scale convolution (kernel widths $2$--$6$) that captures
local character $n$-grams, and then through $3$ self-attention layers, producing
context-aware character vectors $H=(h_1,\dots,h_L)\in\mathbb{R}^{L\times d}$. A
defining property of Turkish is that morpheme identity is governed by
\emph{position relative to the root}: suffixes attach in a fixed slot order
(number, then possessive, then case), so the same surface syllable plays a
different role depending on how many morphemes precede it. In
\textit{ev\,|\,ler\,|\,imiz\,|\,de} (``in our houses''), \textit{-ler} is plural
in the first post-root slot, \textit{-imiz} first-person-plural possessive in
the second, and \textit{-de} locative in the third. The model must therefore
reason about \emph{offsets between characters}---how far a candidate boundary is
from the previous one---rather than their absolute indices. For this reason both
the character encoder and the boundary detector apply Rotary Position Embedding
(RoPE) \citep{su2021rope} on each attention head's subspace, injecting
\emph{relative} offsets directly into the attention dot-product so that a single
learned pattern (e.g.\ ``two characters past the previous boundary'')
generalizes across roots of different lengths.

\paragraph{Boundary detector.}
A stack of $4$ RoPE self-attention layers over $H$, followed by an adjacent-pair
scoring head, emits for each inter-character position a boundary probability
\begin{equation}
    p_i \;=\; \sigma\!\big(\mathrm{score}(h_i, h_{i+1})\big) \in [0,1]
\end{equation}
for each inter-character position $i=1,\dots,L-1$. The vector
$\mathbf{p}=(p_1,\dots,p_{L-1})$ is the only interface to the rest of
the model: everything downstream is a differentiable function of $\mathbf{p}$.

\paragraph{Differentiable Poisson--binomial segmentation.}
The central difficulty is turning soft per-position boundary probabilities into
discrete morpheme segments \emph{without} a non-differentiable
$\arg\max$/threshold that would block gradients from the semantic objectives back
to the boundary detector. It is resolved with a Poisson--binomial dynamic program
that computes, in closed form, the soft assignment of each character to each
segment. Let $b_i\in\{0,1\}$ be the latent boundary indicator at position $i$
with $\Pr[b_i\!=\!1]=p_i$, taken independent. Character $j$ belongs to segment
$k$ (0-indexed) exactly when $k$ boundaries occur before it, i.e.\
$\sum_{i<j} b_i = k$. Since the $p_i$ differ, $\sum_{i<j} b_i$ follows a
\emph{Poisson--binomial} distribution, whose mass is accumulated by
\begin{equation}
    f_j[k] \;=\; f_{j-1}[k]\,(1-p_{j-1}) \;+\; f_{j-1}[k-1]\,p_{j-1},
    \label{eq:pb}
\end{equation}
with base case $f_1[0]=1$ and $f_j[k]=\Pr[\sum_{i<j}b_i=k]$. The resulting matrix $M[j,k]=f_j[k]\in
\mathbb{R}^{L\times S}$ (with $S$ the maximum number of segments and
$\sum_k M[j,k]=1$) is a \emph{soft segment-membership} matrix: row $j$ is a
distribution over which morpheme character $j$ belongs to.
Equation~\eqref{eq:pb} is differentiable in $\mathbf{p}$, costs $O(LS)$, and has
three properties exploited by design. \textbf{(i) Differentiability:} gradients
from the word-level objectives flow through $M$ into the boundary detector, so
boundaries are shaped both by the teacher and by what produces good embeddings.
\textbf{(ii) Soft/hard duality:} as $p_i\!\to\!\{0,1\}$ each row of $M$ converges
to one-hot, recovering exact hard segmentation; the same module yields soft
memberships in training and discrete morphemes at inference, switched only by the
training flag. \textbf{(iii) Surface preservation:} $M$ only \emph{groups}
characters---it never inserts, drops, or rewrites them---so concatenating the
segments reproduces the input word, which is why
$\mathrm{decode}(\mathrm{encode}(w))=w$ holds by construction.

\paragraph{Segment pooling and the word embedding.}
Each segment $k$ is summarized by attention-pooling the character vectors
weighted by their membership, $s_k=\sum_j \alpha_{jk}h_j$ with
$\alpha_{jk}\propto M[j,k]\exp(a(h_j))$ for a learned scorer $a(\cdot)$, so that
within-segment characters compete while cross-segment leakage is suppressed by
$M$. The word embedding is the mean of the valid segment vectors followed by a
two-layer feed-forward network with residual LayerNorm,
$e_w=\mathrm{LayerNorm}(\mathrm{FFN}(\frac{1}{S'}\sum_k s_k))\in\mathbb{R}^{320}$.
Because $e_w$ comes from the same forward pass that yields the boundaries, the
morpheme structure that defines the tokenization is exactly the structure pooled
into the embedding---the architectural basis for treating Morpheus as a
tokenizer and an embedder at once.

\subsection{From segments to token ids}
\label{sec:ids}
Property (iii) makes segmentation lossless for a single word; the tokenizer
extends this to arbitrary text with three steps, each of which has an exact
inverse. \textbf{Pre-tokenization} splits text into word runs, single
punctuation or symbol characters, and whitespace, discarding nothing. One space
before a unit is carried by a SentencePiece-style word-boundary marker (U+2581)
on its first token; every additional space is an explicit marker token, and newline, tab, and
carriage return have dedicated tokens, so runs of spaces and indentation survive.
Word runs longer than the model's $30$-character window are cut into $30$-character
chunks that are segmented independently. \textbf{Segmentation} applies Morpheus
to each word chunk. \textbf{Id mapping} looks each segment up in the vocabulary
(with the boundary marker when it starts a word); a segment that is missing is
emitted as the marker followed by the bare segment, then as single characters,
and any character outside the vocabulary is emitted as its UTF-8 bytes through
$256$ reserved byte tokens. Decoding concatenates tokens, reassembles byte runs
into characters, and turns markers back into spaces. No step normalizes case,
Unicode form, or whitespace, so $\mathrm{decode}(\mathrm{encode}(x))=x$ for any
input text, and no input maps to an unknown token. The $50$K vocabulary is
distilled from Morpheus's own hard segmentations of the training corpus, counting
word-initial and word-internal segments and spaced punctuation exactly as the
tokenizer emits them; the byte, whitespace, and symbol tokens are reserved inside
this budget.

\subsection{Training}
The total loss is a weighted sum of four terms,
\begin{equation}
    \mathcal{L} = w_{\text{aux}}\mathcal{L}_{\text{aux}}
    + w_{\text{sgns}}\mathcal{L}_{\text{sgns}}
    + w_{\text{ctr}}\mathcal{L}_{\text{ctr}}
    + w_{\text{mlm}}\mathcal{L}_{\text{mlm}}.
\end{equation}
$\mathcal{L}_{\text{aux}}$ is a deep-supervised boundary BCE plus a count
regularizer against the (root-corrected) Morfessor labels; its weight follows a
curriculum, decaying geometrically from $0.50$ to $0.08$ over $10$ epochs so the
teacher anchors early training and then yields to the distributional signals.
$\mathcal{L}_{\text{sgns}}$ is skip-gram negative sampling ($16$ negatives,
$\pm6$ window, $120$K context vocabulary); $\mathcal{L}_{\text{ctr}}$ is an
InfoNCE contrastive loss on root identity (the Morfessor first segment,
temperature $0.10$); and $\mathcal{L}_{\text{mlm}}$ is a vocabulary-free
character-level reconstruction in which $20\%$ of words in a sentence are masked
and regenerated character-by-character by a small encoder--decoder. Optimization uses
AdamW, a cosine learning-rate schedule, and gradient clipping, using an
effective batch of $512$ (batch $256\times$ gradient accumulation $2$) for $10$
epochs. TF32 matmuls are enabled while loss components are computed in FP32 for
numerical stability; AMP/BF16 is left off for reproducibility. Training runs in
roughly $30$ minutes per epoch ($\sim$5 hours total) on a single NVIDIA A100
$80$GB. Training dynamics---loss convergence, the per-objective curves, the
aux-weight curriculum, and optimization stability---are reported in
Section~\ref{sec:training-dynamics}.

\section{Results}

\subsection{Training dynamics}
\label{sec:training-dynamics}
\begin{figure*}[t]
    \centering
    \includegraphics[height=3.7cm]{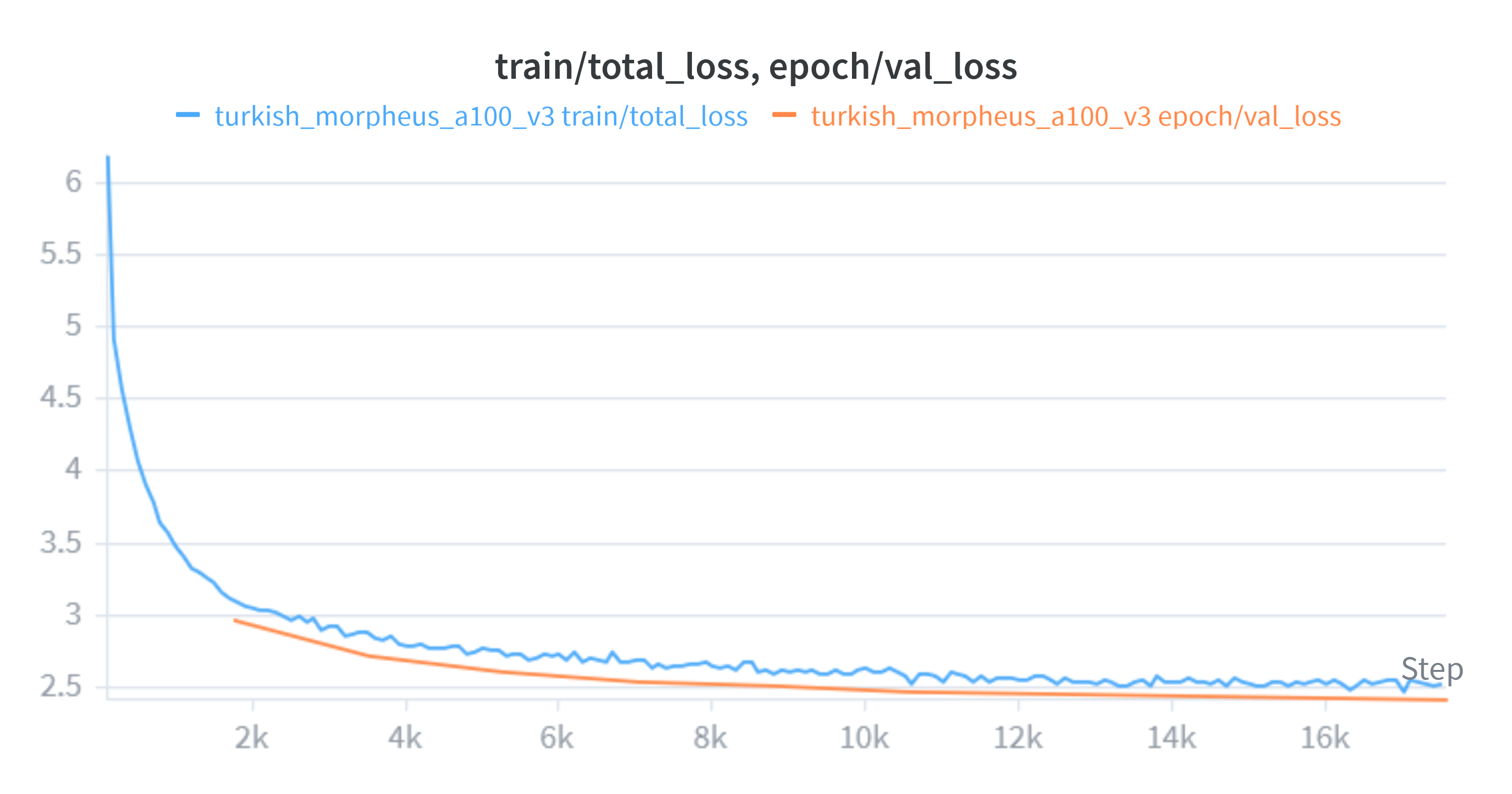}\hfill
    \includegraphics[height=3.7cm]{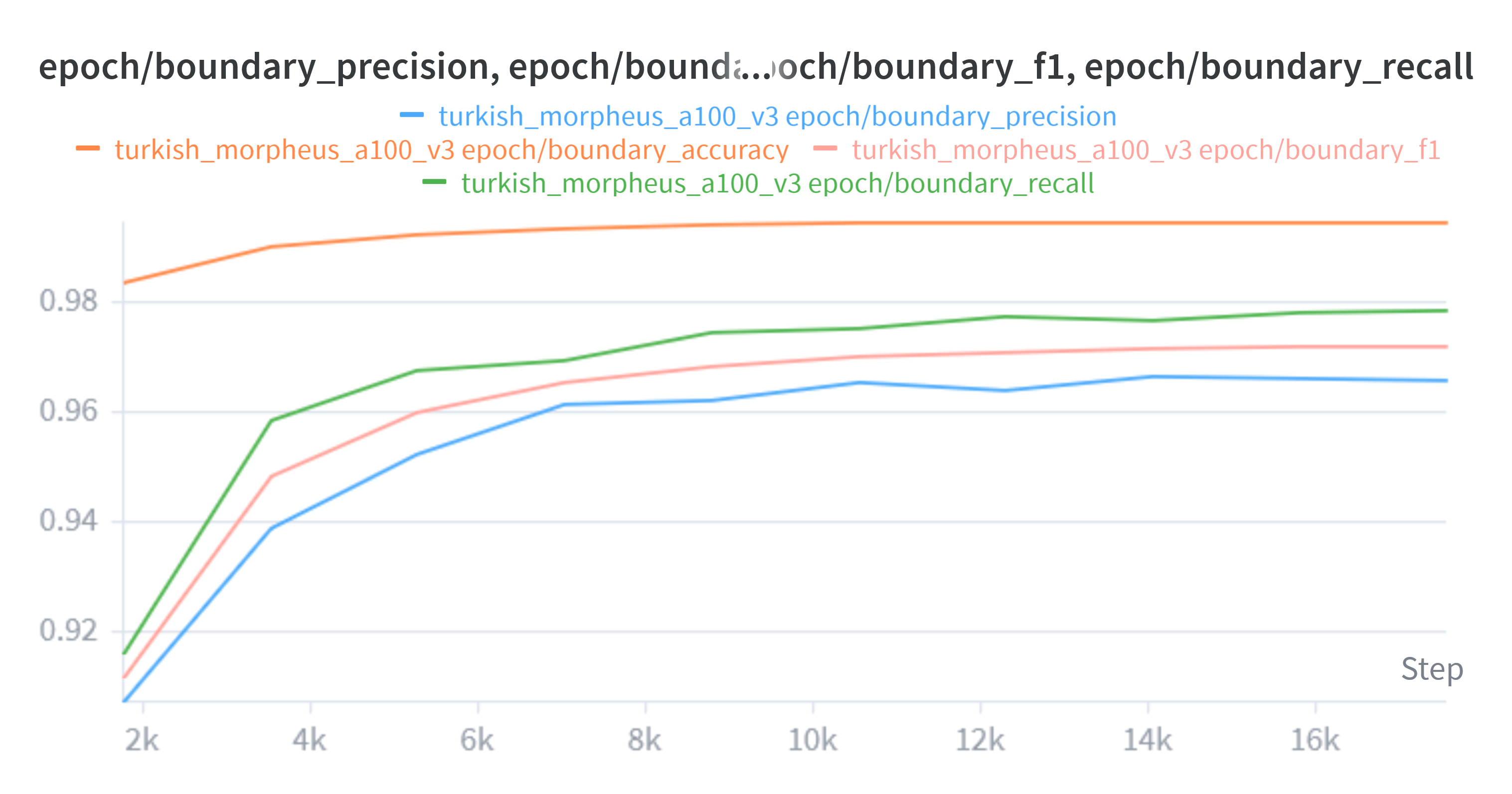}
    \caption{Training dynamics. Left: total train/validation loss. Right:
    boundary-detection precision, recall, F1, and accuracy over training.}
    \label{fig:training}
\end{figure*}

\begin{figure*}[t]
    \centering
    \includegraphics[height=3.4cm]{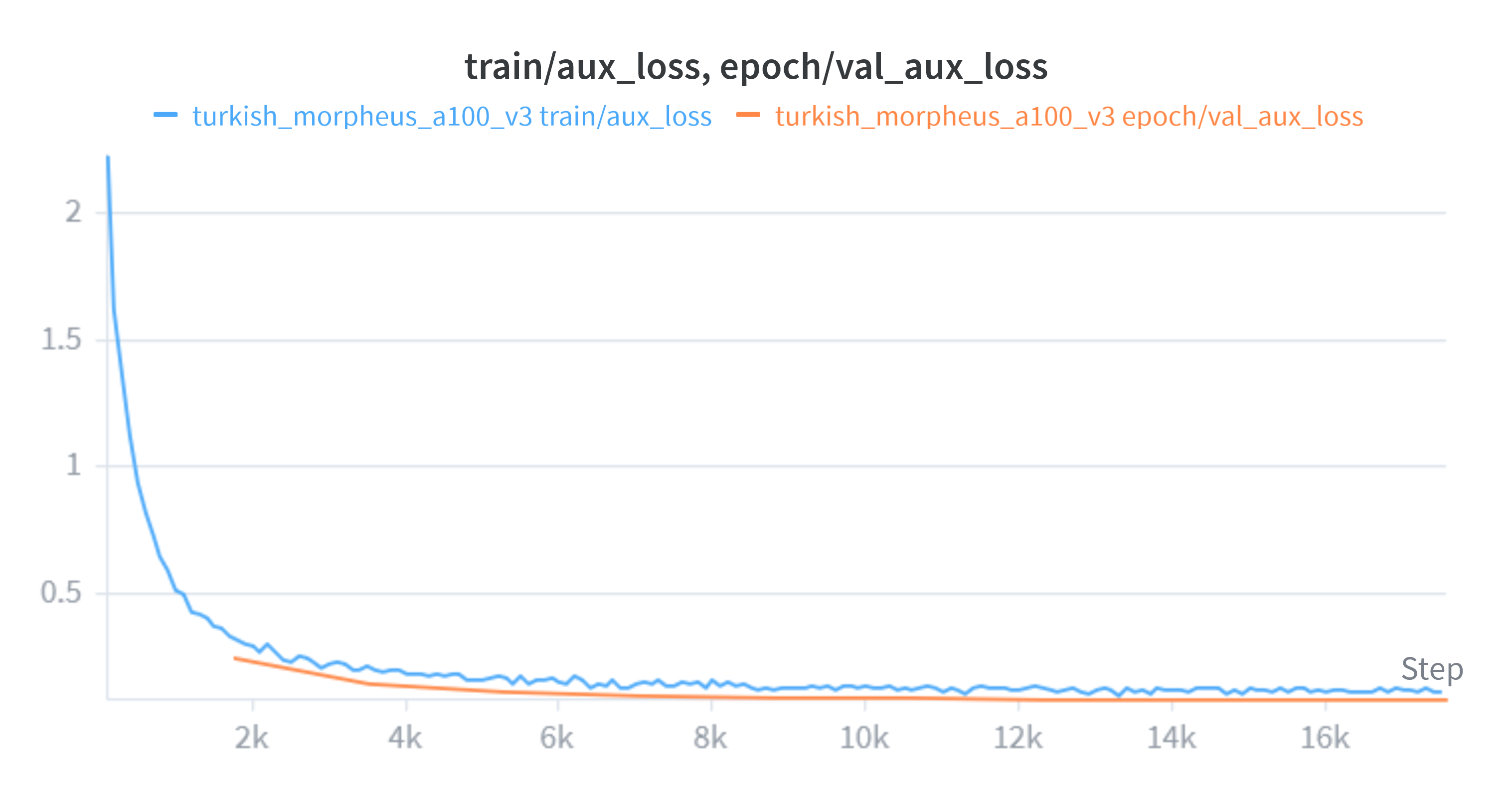}\hfill
    \includegraphics[height=3.4cm]{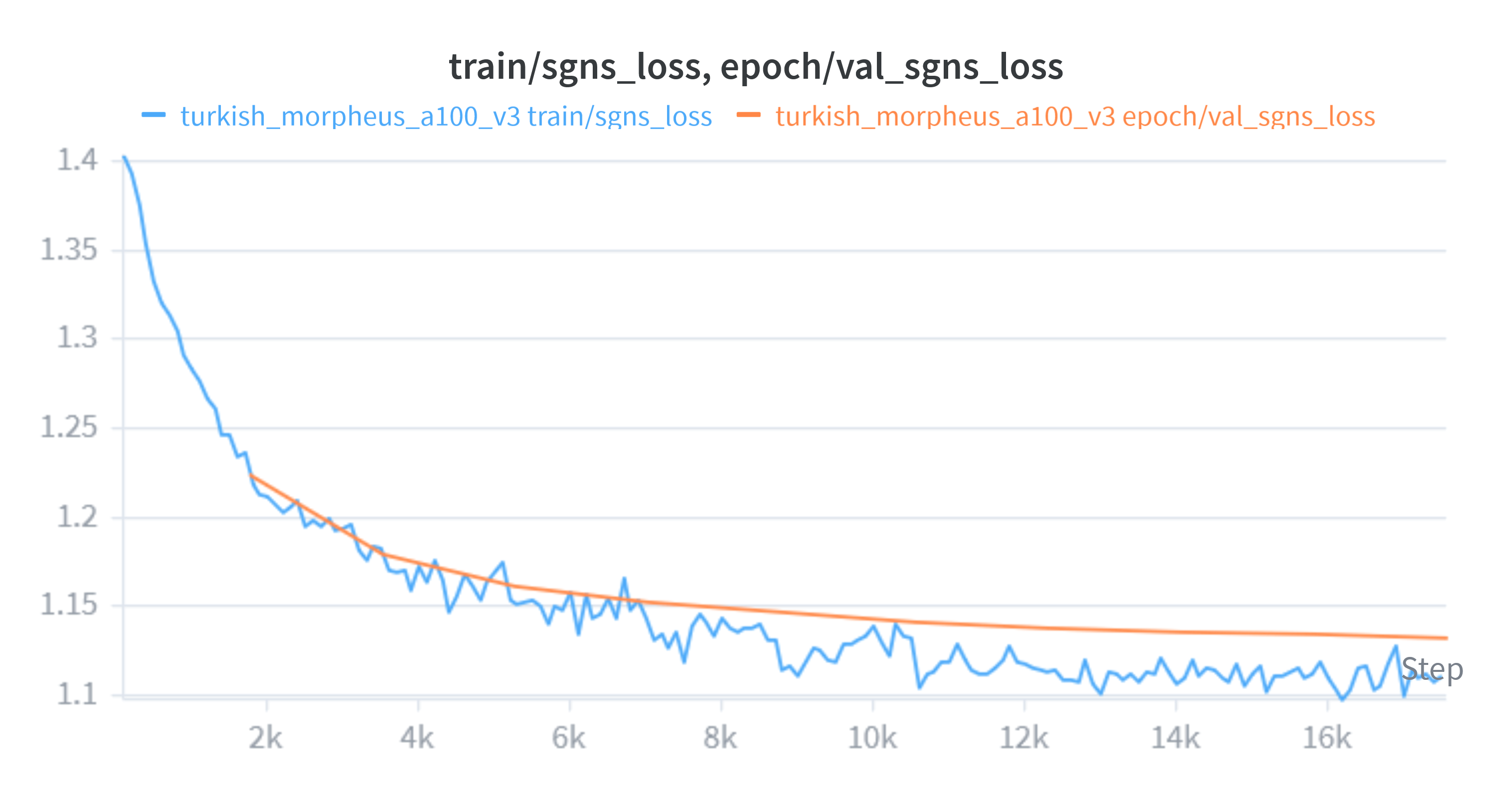}\\[6pt]
    \includegraphics[height=3.4cm]{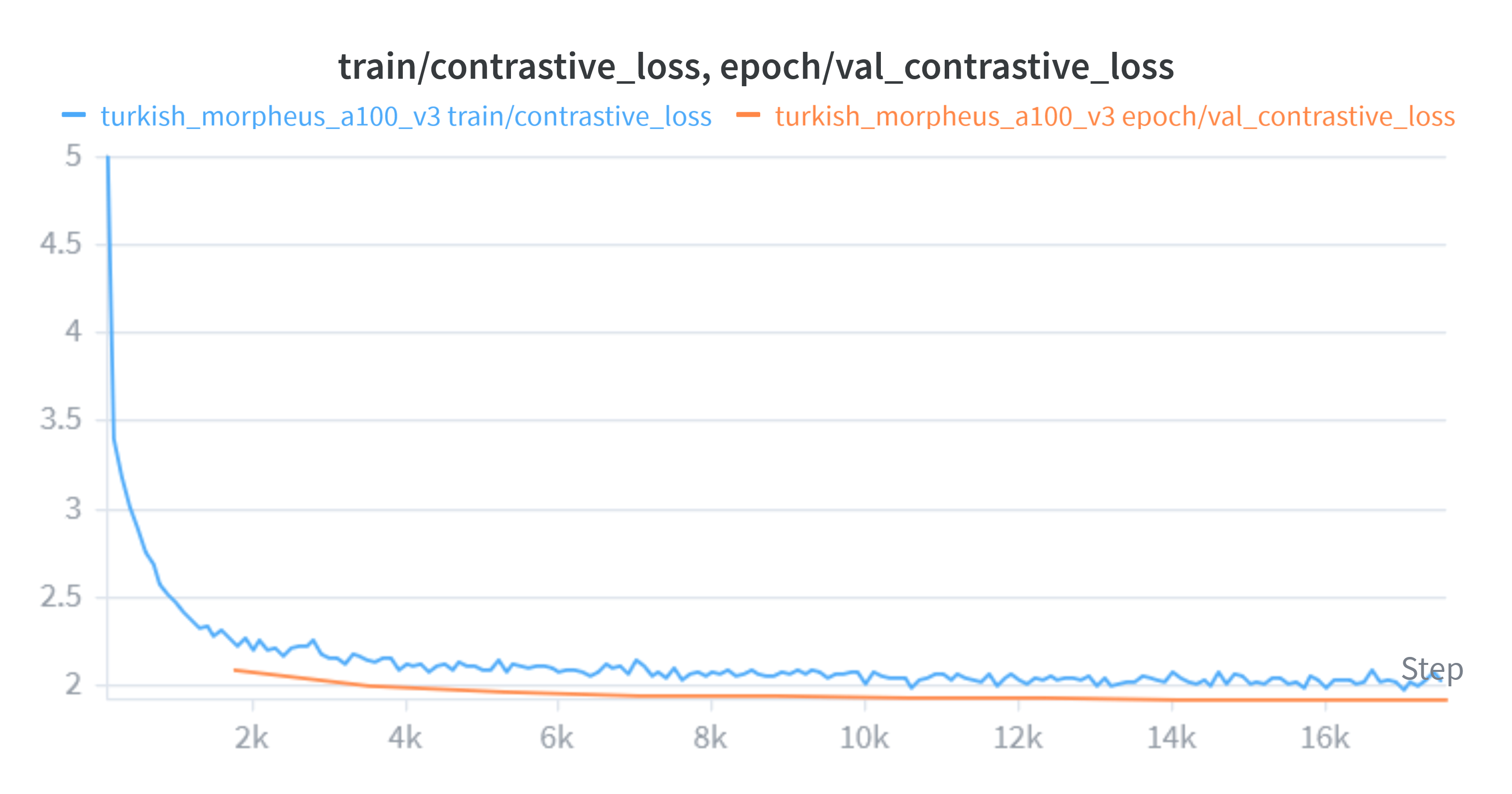}\hfill
    \includegraphics[height=3.4cm]{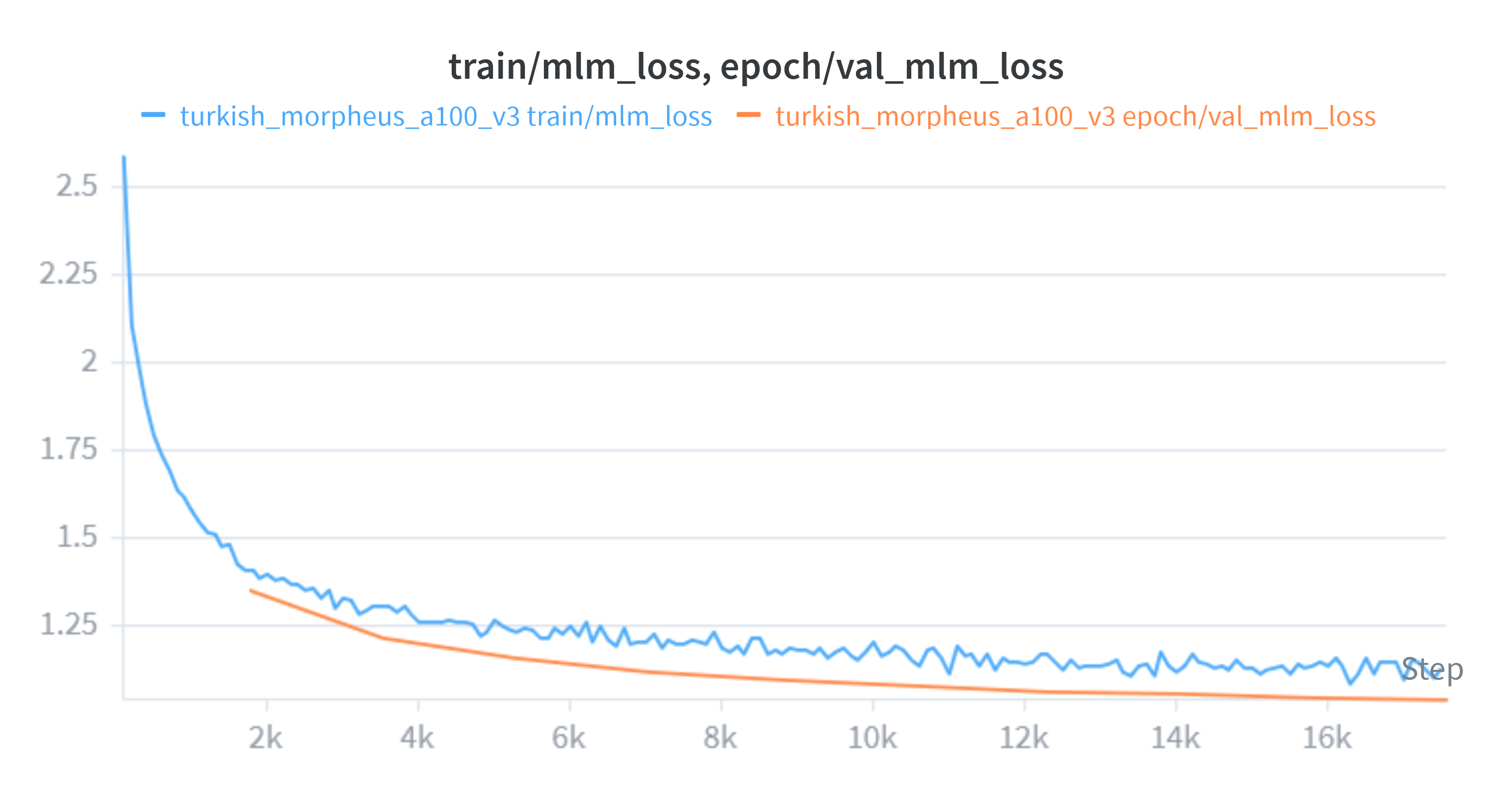}
    \caption{Per-objective train/validation curves: auxiliary boundary loss,
    skip-gram (SGNS), root-identity contrastive, and character-level MLM.}
    \label{fig:losses}
\end{figure*}

\begin{figure*}[t]
    \centering
    \includegraphics[height=3.2cm]{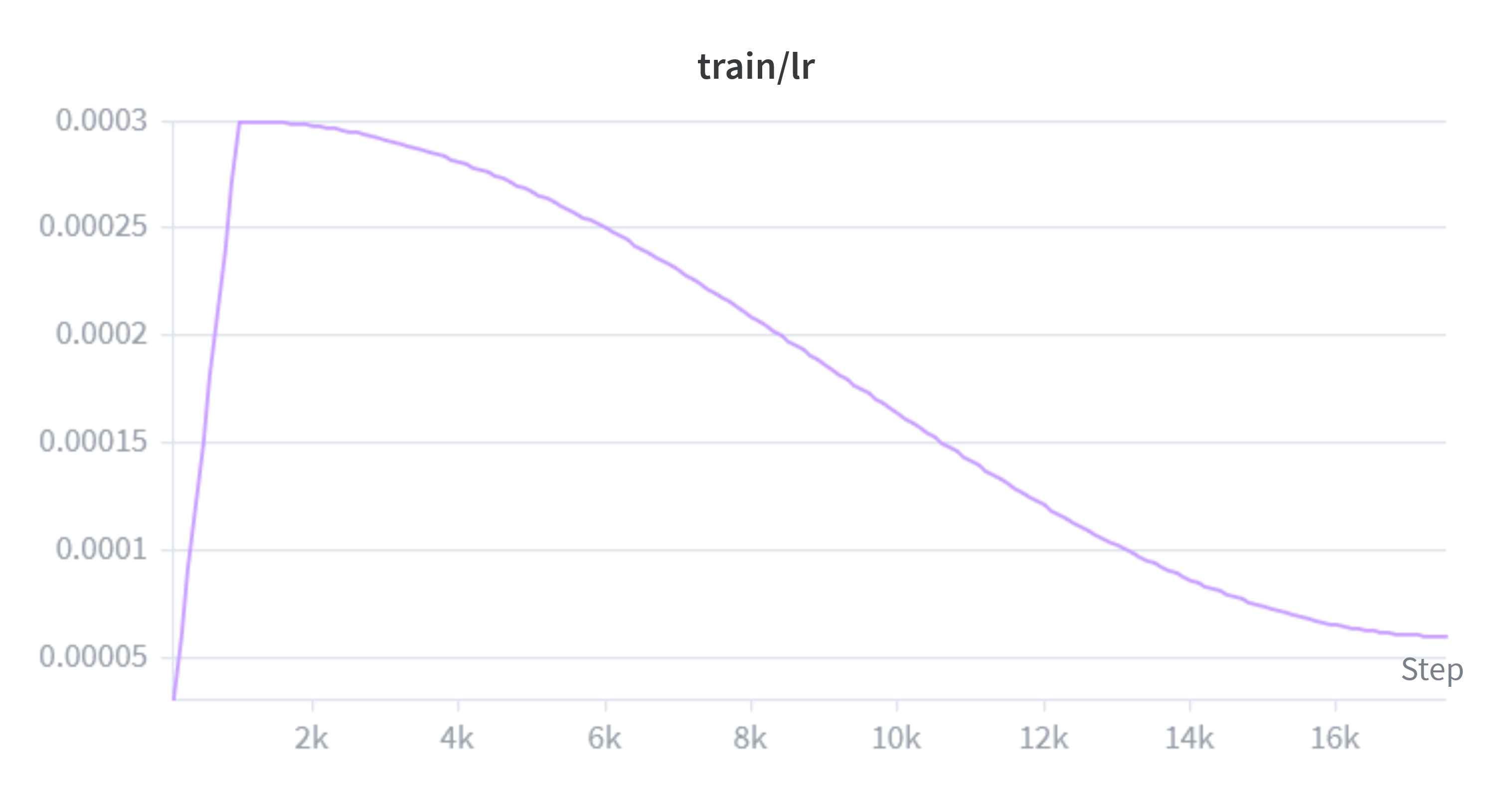}\hfill
    \includegraphics[height=3.2cm]{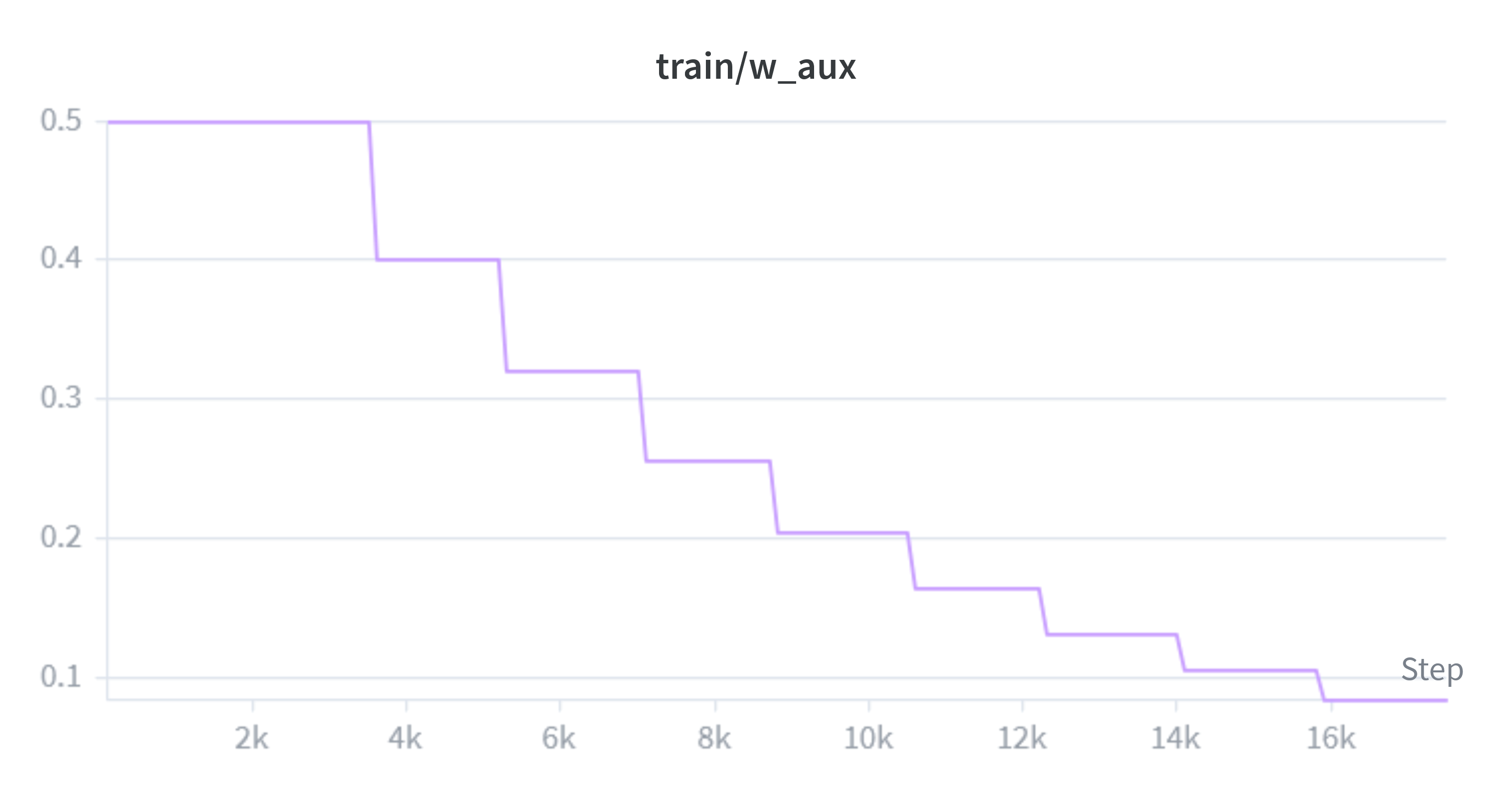}\hfill
    \includegraphics[height=3.2cm]{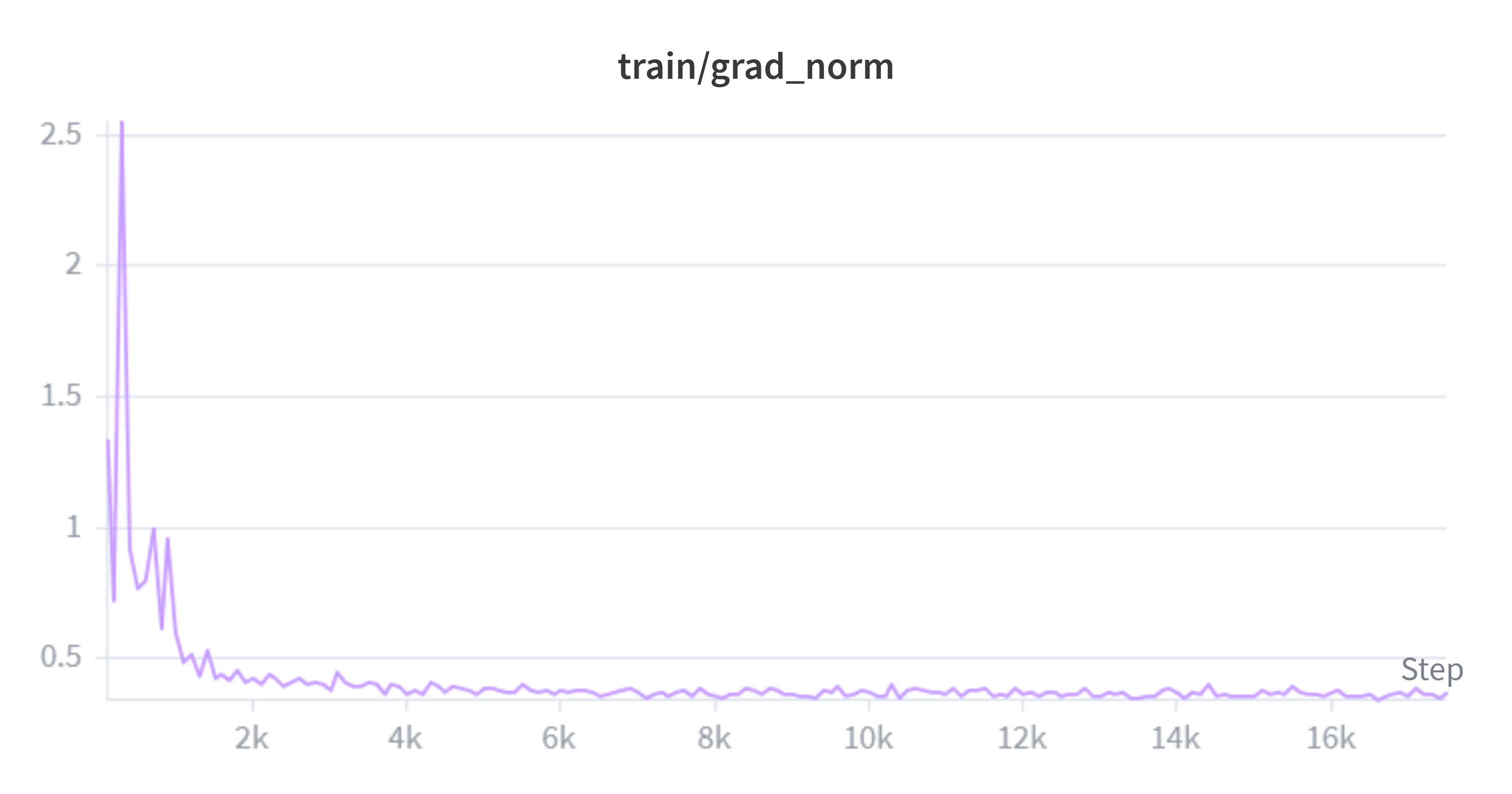}
    \caption{Optimization regime. Left: cosine learning-rate schedule. Middle:
    geometric decay of the auxiliary-loss weight ($0.50\!\rightarrow\!0.08$),
    realizing the teacher-to-distributional curriculum. Right: gradient norm,
    stable throughout under full-precision training.}
    \label{fig:optim}
\end{figure*}

Figure~\ref{fig:training} shows that the total train and validation loss decrease
smoothly and track each other without divergence, while the boundary detector's
precision, recall, and F1 rise quickly and then plateau---confirming that the
Morfessor-supervised objective is learned early. The four objectives converge
jointly (Figure~\ref{fig:losses}): the auxiliary boundary loss drops fastest as
the teacher anchors the early epochs, while the skip-gram, contrastive, and MLM
losses continue to shape the embedding geometry afterwards. Figure~\ref{fig:optim}
documents the optimization regime behind these curves: the cosine learning-rate
schedule, the geometric decay of the auxiliary weight from $0.50$ to $0.08$ that
realizes the teacher-to-distributional curriculum, and a gradient norm that stays
bounded throughout---evidence that running in full precision (AMP off) yields a
stable, reproducible trajectory.

\subsection{Experimental setup}
\label{sec:setup}
\paragraph{Baselines.}
All subword baselines are trained on the same rebuilt training split with a
$64$K vocabulary: BPE and Unigram with SentencePiece \citep{kudo2018sentencepiece}
(\texttt{nmt\_nfkc} normalization, case preserved, byte fallback); WordPiece with
HuggingFace \texttt{tokenizers}, cased and accent-preserving (NFC only, as in
BERTurk-cased); and a true byte-level BPE in the GPT-2 style
\citep{radford2019gpt2}, with no normalizer and byte-level pre-tokenization and
decoding. The comparison also includes a Morfessor Baseline model \citep{creutz2007morfessor}
retrained on the rebuilt split (distinct from the earlier Morfessor model that
supplied Morpheus's training labels) and the rule-based TurkishTokenizer
\citep{bayram2025twm}. None of the tokenizers folds case, so every language model
below predicts the same cased text. Morpheus uses its $50$K vocabulary
(Section~\ref{sec:ids}) and applies no normalization.

\paragraph{Language-modeling protocol.}
Each tokenizer trains a parameter-equalized GPT of $\sim$58M parameters ($8$
layers, sequence length $512$, batch $32$) for exactly $10{,}000$ optimizer steps
with the same cosine schedule. Equal steps means an equal number of
\emph{tokens} processed ($164$M), i.e.\ equal compute; tokenizers that emit more
tokens per character therefore see less text (Section~\ref{sec:lm}). Training uses
the first $1$M lines of the training split; the next $20$K lines form a
validation split used for all intermediate evaluations; the held-out test split
($100$K lines, $23.9$M characters) is scored once, on the final weights. BPC is
the summed test negative log-likelihood in bits divided by the test character
count, which makes it comparable across tokenizers with different
granularities. Each configuration was run with a single seed; the implications are
discussed in Section~\ref{sec:limits}.

\paragraph{Evaluation data.}
Word-level reversibility and MorphScore use $30$K inflected wordforms from
UD\_Turkish-Kenet; inflectional alignment uses the SIGMORPHON~2022 Turkish set
($856$ single-word forms); text-level reversibility uses the first $5{,}000$ raw
lines of the held-out split and a $14$-line stress set covering punctuation,
casing, numbers, URLs, indented code, tabs and CRLF line endings, irregular
spacing, non-Turkish letters, emoji, and characters outside every vocabulary;
surface fidelity uses $50$ curated words grouped by morphological phenomenon;
token validity uses TR-MMLU \citep{bayram2025standards}. Embedding evaluations use
frozen word vectors from Morpheus, BERTurk \citep{schweter2020berturk}, and
BGE-M3 \citep{chen2024bgem3}.

\subsection{Reversibility}
\label{sec:reversibility}
Table~\ref{tab:roundtrip} reports exact reconstruction,
$\mathrm{decode}(\mathrm{encode}(x))=x$, at three levels of difficulty. On
isolated lowercase wordforms every tokenizer except TurkishTokenizer is
reversible; TurkishTokenizer recovers $95.4\%$ because its canonical
re-harmonization rewrites surface suffixes---it maps \textit{saatlerde} (``at
the hours'') to \textit{saat\,$|$\,lar\,$|$\,da}, which decodes to the non-word
\textit{saatlarda}.

\begin{figure}[t]
    \centering
    \includegraphics[width=\linewidth]{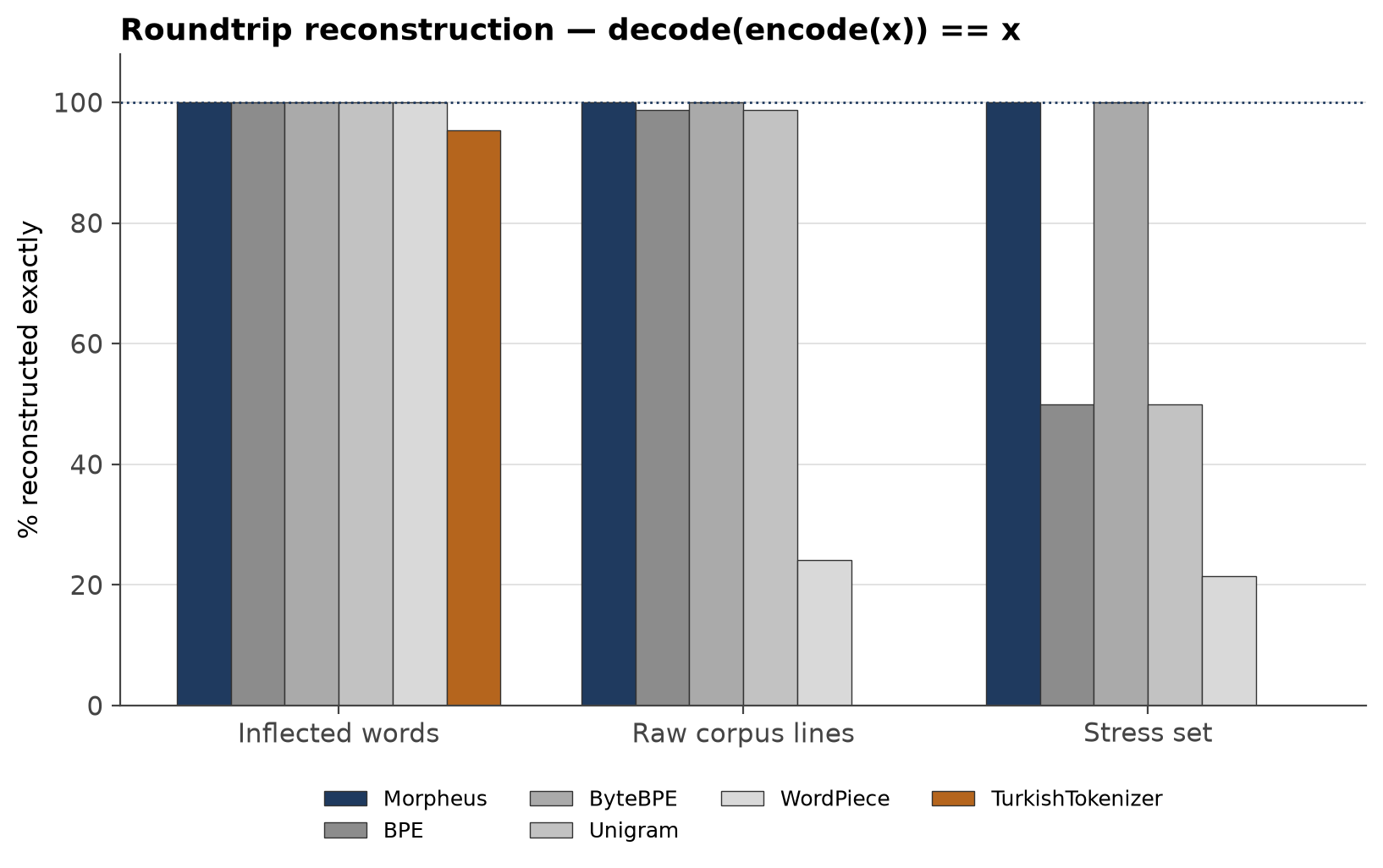}
    \caption{Roundtrip accuracy by level. On isolated words all but
    TurkishTokenizer are reversible; on raw text and the stress set only Morpheus
    and byte-level BPE reconstruct every input.}
    \label{fig:roundtrip}
\end{figure}

Isolated words hide most of what a tokenizer does to running text. On $5{,}000$
raw corpus lines only Morpheus and byte-level BPE reconstruct every input. BPE
and Unigram reach $98.7\%$; their failures come from NFKC normalization
(superscript $^2\!\to\!2$, the ellipsis character$\,\to\,$three periods), collapsed
runs of spaces, and dropped non-breaking or zero-width characters. WordPiece
falls to $24.2\%$ because its whitespace pre-tokenization cannot restore whether
punctuation was attached to the preceding word, and TurkishTokenizer to $0\%$,
since it both rewrites suffixes and inserts spaces between pieces. The stress set
sharpens the picture: Morpheus and byte-level BPE pass all $14$ lines, BPE and
Unigram $7$ (they also lose newlines, tabs, and indentation), WordPiece $3$, and
TurkishTokenizer none (Figure~\ref{fig:roundtrip}).

Two conclusions follow. First, the failures of BPE, Unigram, and WordPiece are
properties of their preprocessing rather than of their segmentation algorithms;
a SentencePiece model trained with identity normalization would avoid most of
them. Second, exact reversibility is achieved by two very different designs:
byte-level BPE guarantees it by never leaving the byte level, Morpheus by
segmenting only surface strings and falling back to bytes for unseen characters
(Section~\ref{sec:ids}). Reversibility therefore does not distinguish Morpheus
from the strongest subword baseline; the remaining sections ask what does.

\begin{table}[t]
\centering\small
\setlength{\tabcolsep}{4pt}
\begin{tabular}{lccc}
\toprule
 & Words & Raw text & Stress \\
Tokenizer & ($30{,}204$) & ($5{,}000$) & ($14$) \\
\midrule
\textbf{Morpheus} & \textbf{100.0} & \textbf{100.0} & \textbf{14/14} \\
ByteBPE & \textbf{100.0} & \textbf{100.0} & \textbf{14/14} \\
BPE & \textbf{100.0} & 98.7 & 7/14 \\
Unigram & \textbf{100.0} & 98.7 & 7/14 \\
WordPiece & \textbf{100.0}$^{*}$ & 24.2 & 3/14 \\
TurkishTokenizer & 95.4 & 0.0 & 0/14 \\
\bottomrule
\end{tabular}
\caption{Exact reconstruction (\%) of isolated inflected words, raw held-out
corpus lines, and a stress set. BPE and Unigram lose NFKC-normalized characters
and whitespace; WordPiece loses spacing around punctuation; TurkishTokenizer
canonicalizes allomorphs and inserts spaces. $^{*}$One failure in $30{,}204$.}
\label{tab:roundtrip}
\end{table}

\subsection{Surface fidelity}
A tokenizer can place boundaries well yet still corrupt the surface string. This
is probed with $50$ curated OOV-leaning Turkish words, scoring each
segmentation along four increasingly strict criteria (Table~\ref{tab:qualitative}):
\emph{root\%}, whether the first segment is the correct root; \emph{count\%},
whether the number of segments matches the gold; \emph{len\%}, whether the
segment \emph{lengths} match (i.e.\ the boundaries are placed correctly); and
\emph{exact\%}, whether the segment \emph{strings} exactly match the surface
morphemes.

Morpheus segments $38\%$ of the words exactly, the highest of all tokenizers
(Morfessor $28\%$, the subword family $6$--$22\%$), and identifies the root most
often of all tokenizers ($66\%$). Its len\% and exact\% coincide: every
boundary it places is also a faithful surface string. TurkishTokenizer shows the
opposite pattern. It places boundaries best ($\text{count}=92\%$,
$\text{len}=78\%$), but its strings match the surface only $10\%$ of the time, a
$68$-point gap that isolates its canonicalization: on \textit{saatlerde},
\textit{rollerde}, and \textit{harflerle} it returns \textit{saat\,$|$\,lar\,$|$\,da},
\textit{rol\,$|$\,lar\,$|$\,da}, and \textit{harf\,$|$\,lar\,$|$\,la}---correct
boundaries with rewritten suffixes. The subword tokenizers do not rewrite these
words, so their len\% equals their exact\%, but they rarely align with morphology;
byte-level BPE is lowest ($6\%$): it often produces the right number of segments
(count $48\%$) but almost never at the right positions.
Table~\ref{tab:decode} traces three words through each tokenizer: all reversible
tokenizers reproduce the input, and they differ only in where they cut.

\begin{table}[t]
\centering\small
\setlength{\tabcolsep}{4pt}
\begin{tabular}{lcccc}
\toprule
Tokenizer & root\% & count\% & len\% & exact\% \\
\midrule
\textbf{Morpheus} & \textbf{66} & 46 & 38 & \textbf{38} \\
Morfessor & 58 & 44 & 28 & 28 \\
WordPiece & 32 & 32 & 22 & 22 \\
Unigram & 30 & 28 & 18 & 18 \\
BPE & 32 & 28 & 16 & 16 \\
ByteBPE & 24 & 48 & 6 & 6 \\
\midrule
TurkishTok.$^\dagger$ & 64 & \textbf{92} & \textbf{78} & 10 \\
\bottomrule
\end{tabular}
\caption{Surface fidelity on $50$ curated OOV-leaning words. \emph{root\%}: first
segment is the correct root; \emph{count\%}: segment count matches gold;
\emph{len\%}: boundaries placed correctly; \emph{exact\%}: segment strings match
the surface morphemes. The len\%$\rightarrow$exact\% drop isolates surface
rewriting: zero for Morpheus, $68$ points for TurkishTokenizer.
$^\dagger$Not reversible.}
\label{tab:qualitative}
\end{table}

\paragraph{Where the advantage comes from.}
Breaking the $50$ words down by phenomenon (Table~\ref{tab:qualcat}) shows that
Morpheus's lead is concentrated. On rare and technical roots
(\textit{katalizörler}, \textit{kromozomlar}, \textit{integraller}) it segments
$70\%$ of words exactly, against $45\%$ for the best subword tokenizer and $40\%$
for Morfessor: the neural boundary model generalizes to roots that statistical
segmenters fragment. On derivational chains Morfessor is better ($40\%$ vs.\
$30\%$), because Morpheus tends to keep derivational suffix sequences together
(\textit{demir\,$|$\,cilik}, \textit{orman\,$|$\,cılık}). Stem alternations are
hard for every surface-preserving tokenizer: with consonant softening
(\textit{yaprağın}) and vowel drop (\textit{burnu}, \textit{aklın}) no tokenizer
exceeds $33\%$, and on loanword harmony exceptions (\textit{rollerde},
\textit{gollerin}) none is exact---Morpheus finds the root but merges the suffix
chain (\textit{rol\,$|$\,lerde}), and TurkishTokenizer matches the boundaries only
by rewriting the suffixes. These counts are small ($6$--$20$ words per row) and
should be read as qualitative evidence.

\begin{table}[t]
\centering\small
\setlength{\tabcolsep}{3pt}
\resizebox{\columnwidth}{!}{%
\begin{tabular}{lcccc}
\toprule
Phenomenon ($n$) & Morpheus & Morfessor & Subword & TT$^\dagger$ \\
\midrule
Rare root (20) & \textbf{70}/\textbf{70} & 40/40 & 45/45 & 90/20 \\
Derivation (10) & 30/30 & \textbf{40}/\textbf{40} & 20/20 & 80/10 \\
Softening (7) & \textbf{14}/\textbf{14} & 0/0 & 0/0 & 100/0 \\
Vowel drop (6) & 17/17 & \textbf{33}/\textbf{33} & 0/0 & 0/0 \\
Loanword exc.\ (7) & 0/0 & 0/0 & 0/0 & 86/0 \\
\bottomrule
\end{tabular}}
\caption{Surface fidelity by phenomenon, as len\%\,/\,exact\%. \emph{Subword} is
the best of BPE, ByteBPE, Unigram, and WordPiece in each row; TT is
TurkishTokenizer. $^\dagger$Not reversible.}
\label{tab:qualcat}
\end{table}

\begin{table*}[t]
\centering\small
\begin{tabular}{lllll}
\toprule
Word (gold) & Tokenizer & Segmentation & $\mathrm{decode}(\mathrm{encode}(w))$ & $=w$? \\
\midrule
\multirow{4}{*}{\textit{köpeğim} (köpeğ\,|\,im)}
 & \textbf{Morpheus} & köpeğ \textbar{} im & köpeğim & \checkmark \\
 & TurkishTokenizer & köpek \textbar{} üm & köpeküm & \ding{55} \\
 & ByteBPE & köp \textbar{} eğim & köpeğim & \checkmark \\
 & BPE & köp \textbar{} eğim & köpeğim & \checkmark \\
\midrule
\multirow{4}{*}{\textit{saatlerde} (saat\,|\,ler\,|\,de)}
 & \textbf{Morpheus} & saatler \textbar{} de & saatlerde & \checkmark \\
 & TurkishTokenizer & saat \textbar{} lar \textbar{} da & saatlarda & \ding{55} \\
 & ByteBPE & saat \textbar{} lerde & saatlerde & \checkmark \\
 & BPE & saatlerde & saatlerde & \checkmark \\
\midrule
\multirow{4}{*}{\textit{çiçeğin} (çiçeğ\,|\,in)}
 & \textbf{Morpheus} & çi \textbar{} çe \textbar{} ğin & çiçeğin & \checkmark \\
 & TurkishTokenizer & çiçek \textbar{} ün & çiçekün & \ding{55} \\
 & ByteBPE & çiç \textbar{} eğin & çiçeğin & \checkmark \\
 & BPE & çiç \textbar{} eğin & çiçeğin & \checkmark \\
\bottomrule
\end{tabular}
\caption{Representative decode outcomes. Every reversible tokenizer reproduces the
input; they differ in where they cut. Morpheus is exact on \textit{köpeğim},
merges the suffix chain of \textit{saatlerde}, and over-segments the softened stem
of \textit{çiçeğin}---an error that costs alignment but not reversibility.
TurkishTokenizer rewrites allomorphs (\textit{-üm}, \textit{-lar}/\textit{-da},
\textit{-ün}) and decodes to non-words.}
\label{tab:decode}
\end{table*}

\subsection{Morphological alignment}
\begin{figure*}[t]
    \centering
    \includegraphics[height=4.2cm]{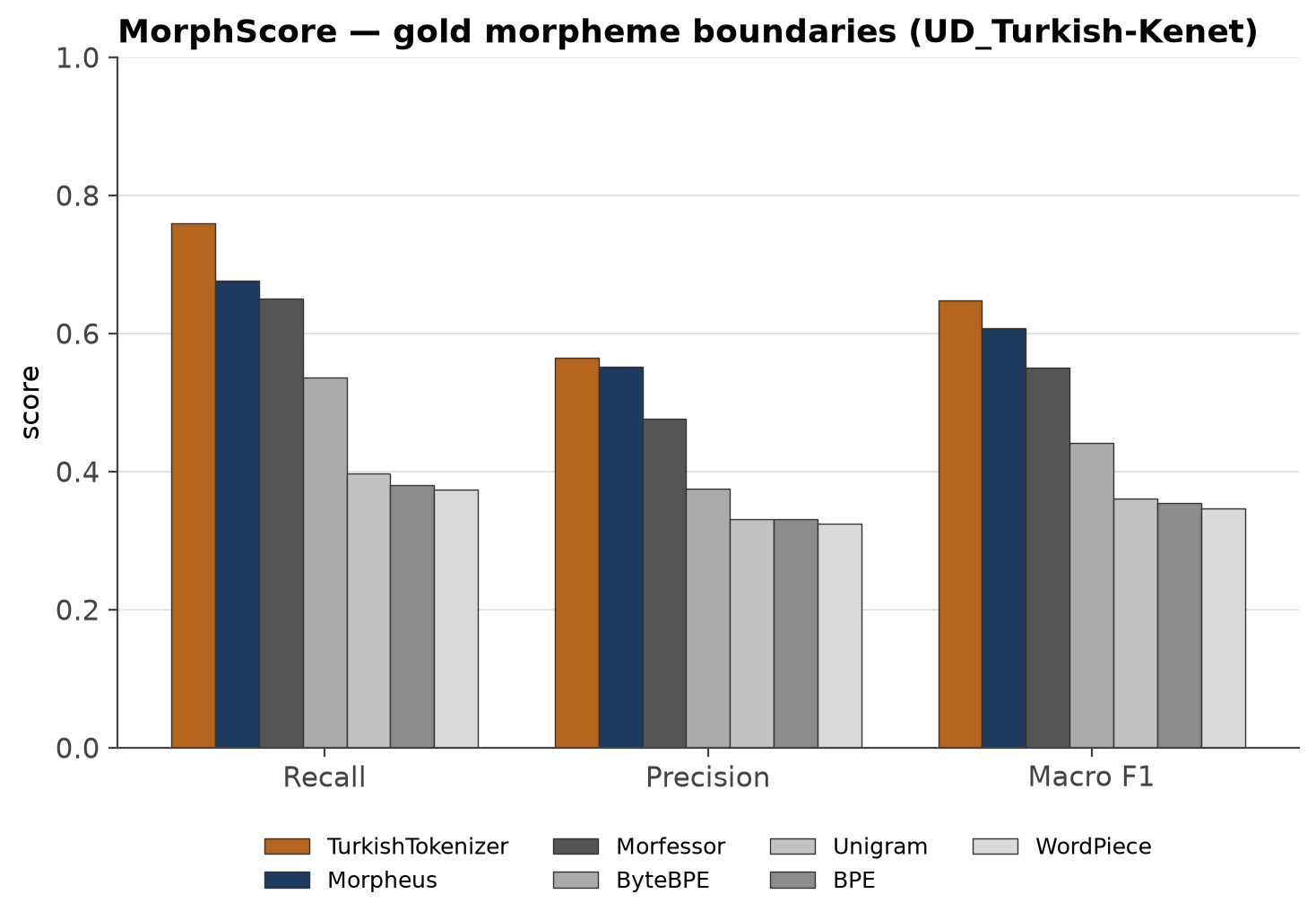}\hfill
    \includegraphics[height=4.2cm]{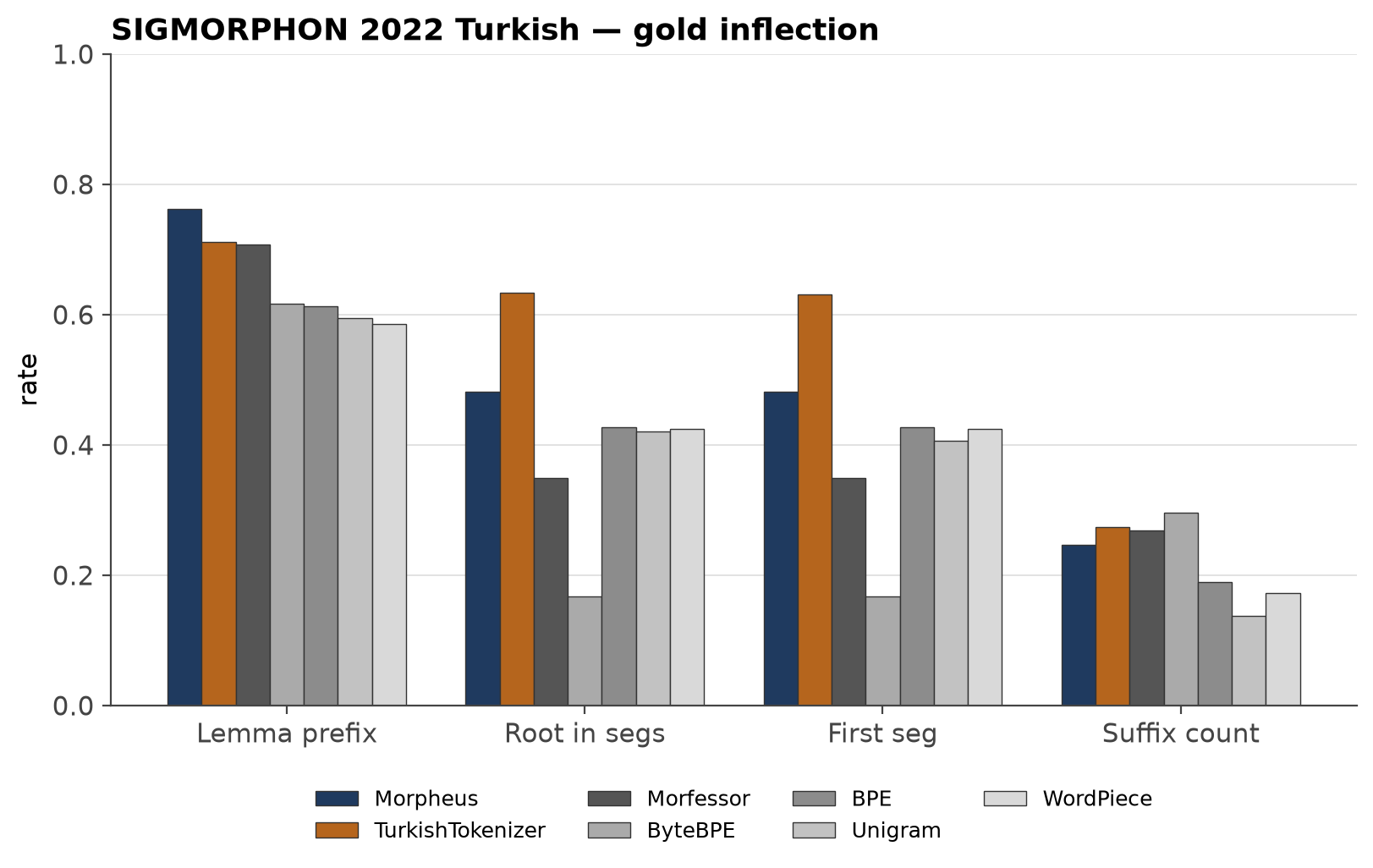}
    \caption{Morphological alignment. Left: MorphScore (UD\_Turkish-Kenet)
    recall, precision, and macro-F1. Right: SIGMORPHON inflection rates.}
    \label{fig:morph}
\end{figure*}

On gold morphological segmentation (Table~\ref{tab:morph},
Figure~\ref{fig:morph}), Morpheus is the strongest reversible tokenizer by a wide
margin. On MorphScore (UD\_Turkish-Kenet) it reaches a macro-F1 of $0.61$, against
$0.44$ for byte-level BPE, $0.35$--$0.36$ for BPE, Unigram, and WordPiece, and
$0.55$ for Morfessor; only the non-reversible TurkishTokenizer scores higher
($0.65$), and it does so with a $1.9\%$ length-mismatch rate caused by
canonicalization, whereas Morpheus's segments always concatenate to the input.
Byte-level BPE's relatively high score comes from recall ($0.54$) rather than
precision ($0.37$): it almost never keeps a word whole (single-token rate $1.4\%$,
against $22$--$27\%$ for the other subword tokenizers), and some of its many
boundaries coincide with morpheme boundaries. On SIGMORPHON inflection,
Morpheus has the highest lemma-prefix rate of all tokenizers ($0.76$; its first
segments most often begin with the lemma), and its root appears as a complete
segment in $48\%$ of forms---more than any other reversible tokenizer
($0.17$--$0.43$) and the retrained Morfessor baseline ($0.35$), though below
TurkishTokenizer's dictionary-driven $0.63$. The root-in-segments rate is
consistent with the root correction applied to Morpheus's training labels
(Section~3.2), but no ablation was run that would isolate its effect.

\begin{table}[t]
\centering\small
\setlength{\tabcolsep}{5pt}
\begin{tabular}{lccc}
\toprule
 & MorphScore & SIGM. & SIGM. \\
Model & macro-F1 & lemma & root \\
\midrule
\textbf{Morpheus} & \textbf{0.61} & \textbf{0.76} & 0.48 \\
Morfessor & 0.55 & 0.71 & 0.35 \\
ByteBPE & 0.44 & 0.62 & 0.17 \\
Unigram & 0.36 & 0.59 & 0.42 \\
BPE & 0.35 & 0.61 & 0.43 \\
WordPiece & 0.35 & 0.59 & 0.42 \\
\midrule
TurkishTok.$^\dagger$ & \textbf{0.65} & 0.71 & \textbf{0.63} \\
\bottomrule
\end{tabular}
\caption{Morphological alignment: MorphScore macro-F1 (UD\_Turkish-Kenet) and
SIGMORPHON lemma-prefix and root-in-segments rates. $^\dagger$Not reversible.}
\label{tab:morph}
\end{table}

\subsection{Token validity on TR-MMLU}
\begin{figure}[t]
    \centering
    \includegraphics[width=\linewidth]{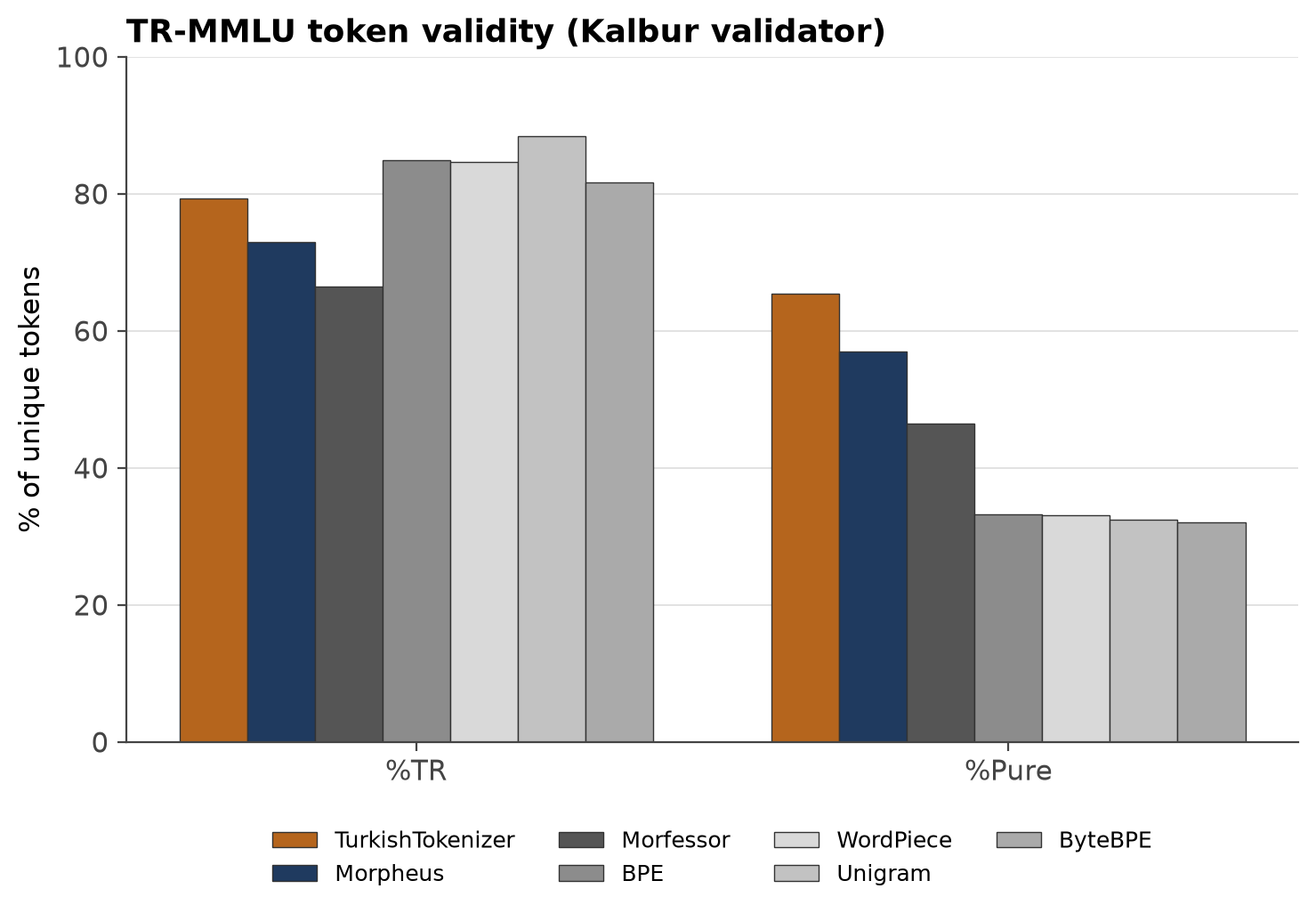}
    \caption{TR-MMLU token validity over unique tokens: Turkish-token (\%TR) and
    pure-token (\%Pure) rates.}
    \label{fig:trmmlu}
\end{figure}

Following \citet{bayram2025standards}, the measure is the share of tokens that are
valid Turkish strings (\%TR) and valid Turkish morphemes or words (\%Pure) on the
TR-MMLU benchmark, both over unique tokens and weighted by frequency
(Table~\ref{tab:trmmlu}, Figure~\ref{fig:trmmlu}). Over unique tokens Morpheus
has the second-highest pure-token rate ($57.1\%$, against $32$--$33\%$ for the
subword family and $46.5\%$ for Morfessor), behind only the non-reversible
TurkishTokenizer ($65.5\%$). Weighted by frequency, the three
morphological tokenizers cluster together ($76$--$78\%$ \%Pure), far above the
subword family ($48$--$51\%$). The same table shows the cost of morpheme-level
units: Morpheus emits $1.90$ tokens per word, against $1.48$--$1.56$ for the
subword family, of which byte-level BPE is the most fertile ($1.56$).

\begin{table}[t]
\centering\small
\setlength{\tabcolsep}{3.5pt}
\begin{tabular}{lccccc}
\toprule
 & \multicolumn{2}{c}{Unique} & \multicolumn{2}{c}{Freq.-weighted} & Fert. \\
Tokenizer & \%TR & \%Pure & \%TR & \%Pure & tok/w \\
\midrule
\textbf{Morpheus} & 73.0 & \textbf{57.1} & 84.0 & 76.5 & 1.90 \\
Morfessor & 66.5 & 46.5 & 84.6 & 75.9 & 1.82 \\
ByteBPE & 81.7 & 32.1 & 78.1 & 50.2 & 1.56 \\
Unigram & \textbf{88.5} & 32.5 & 77.7 & 48.6 & 1.54 \\
BPE & 85.0 & 33.3 & 77.9 & 48.2 & \textbf{1.48} \\
WordPiece & 84.7 & 33.2 & 81.8 & 51.2 & \textbf{1.48} \\
\midrule
TurkishTok.$^\dagger$ & 79.4 & \textbf{65.5} & \textbf{85.5} & \textbf{78.2} & 1.98 \\
\bottomrule
\end{tabular}
\caption{Token validity on TR-MMLU (Kalbur validator) and fertility. $^\dagger$Not
reversible.}
\label{tab:trmmlu}
\end{table}

\subsection{A controlled language-modeling probe}
\label{sec:lm}
\begin{figure*}[t]
    \centering
    \includegraphics[width=0.92\textwidth]{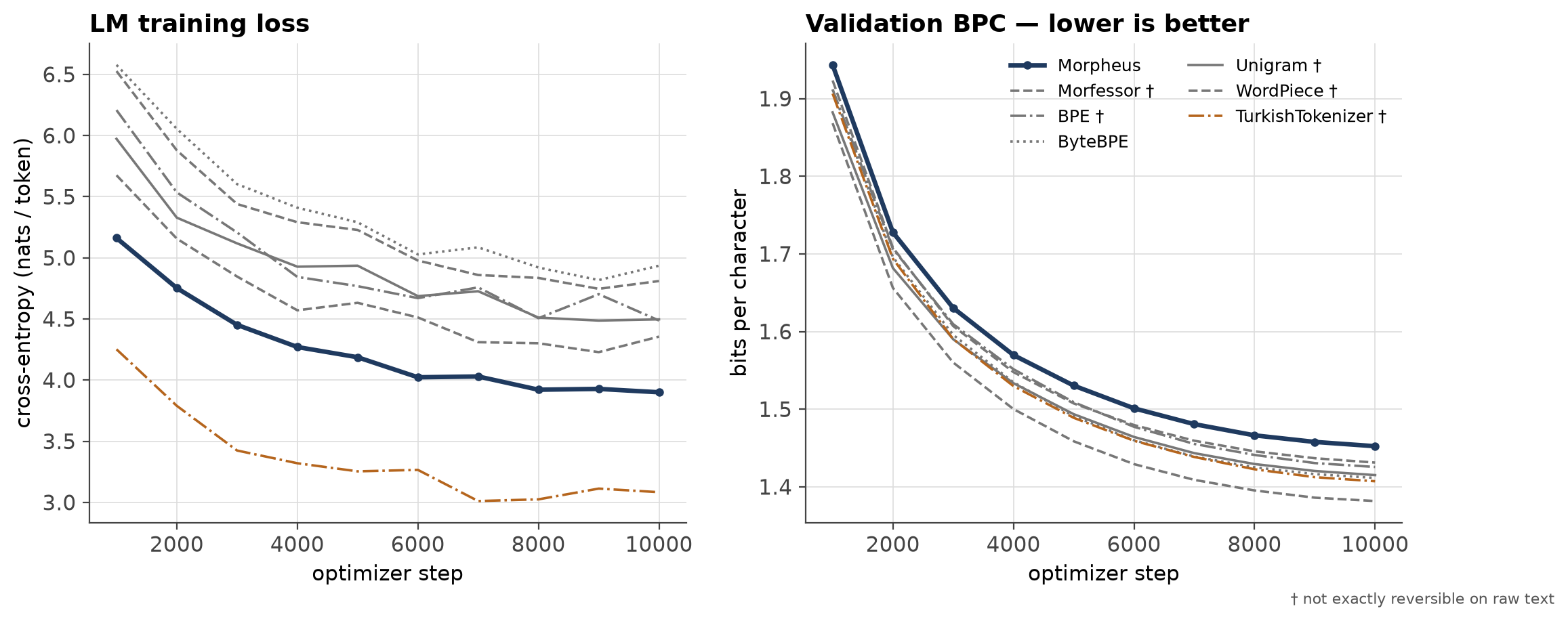}
    \caption{Downstream language-model training at equal compute. Left: training
    loss (nats per token, so not comparable across tokenizers). Right: validation
    BPC on held-out training lines. $\dagger$ marks tokenizers that do not
    reconstruct raw text exactly.}
    \label{fig:lmtrain}
\end{figure*}

Table~\ref{tab:lm} and Figure~\ref{fig:lmtrain} report the language-modeling
results under the equal-compute protocol. Morpheus reaches a test BPC of
$1.451$. Among the two exactly reversible tokenizers, byte-level BPE is better by
$0.040$ BPC ($2.8\%$); the other baselines fall between them (Unigram $1.416$,
BPE $1.425$, Morfessor $1.430$).\footnote{Version~1 reported a BPC of $1.425$ for
Morpheus and ranked it first among reversible tokenizers. That run used a
pre-tokenizer that removed punctuation and words longer than $30$ characters
from Morpheus's token stream while BPC was still normalized by the full
character count, which understated Morpheus's BPC; the baselines also folded
case. The present comparison removes both asymmetries.} The two lowest values
belong to non-reversible tokenizers and are not directly comparable: WordPiece
($1.380$) does not encode whether punctuation is attached to the preceding word,
and TurkishTokenizer ($1.407$) maps allomorphs to shared ids, so both models
predict a less informative token stream than the one their BPC is normalized by.

Two properties of the comparison qualify the gap. First, Morpheus units are
smaller: it averages $3.81$ characters per token, against $4.97$ for byte-level BPE
and $4.60$ for BPE, so the same text is $30\%$ longer as a Morpheus token sequence.
Each Morpheus token is correspondingly cheaper to predict ($3.83$ nats against
$4.86$ for byte-level BPE), and BPC normalizes for this, but the fixed budget of
$164$M tokens covers only $624$M characters of text for Morpheus against $815$M for
byte-level BPE. Equivalently, over the same $1$M training lines Morpheus completes
$2.6$ epochs where byte-level BPE completes $3.4$. Second, the validation curves of
all tokenizers are still descending at $10$K steps, with similar slopes
(Morpheus $-0.0055$ BPC over the last $1$K steps, byte-level BPE $-0.0049$), so the
results describe a fixed budget rather than converged models, and the curves give
no evidence that the gap closes with more training. An equal-text comparison (the
same number of epochs for every tokenizer) and multiple seeds are needed to
separate the effect of tokenization from the effect of the budget definition; this
point is revisited in Section~\ref{sec:assessment}.

\begin{table}[t]
\centering\small
\setlength{\tabcolsep}{3.2pt}
\begin{tabular}{lccccc}
\toprule
 & & Raw & Chars & NLL & GPU \\
Tokenizer & BPC & RT\% & /tok & /tok & MB \\
\midrule
\textbf{Morpheus} & 1.451 & \textbf{100.0} & 3.81 & 3.83 & 3021 \\
ByteBPE & \textbf{1.411} & \textbf{100.0} & 4.97 & 4.86 & 3724 \\
\midrule
Unigram$^\dagger$ & 1.416 & 98.7 & 4.59 & 4.51 & 3724 \\
BPE$^\dagger$ & 1.425 & 98.7 & 4.60 & 4.54 & 3724 \\
Morfessor$^\dagger$ & 1.430 & -- & 4.26 & 4.22 & 3010 \\
WordPiece$^\dagger$ & 1.380 & 24.2 & 4.97 & 4.75 & 3724 \\
TurkishTok.$^\dagger$ & 1.407 & 0.0 & 3.09 & 3.01 & 2152 \\
\bottomrule
\end{tabular}
\caption{Language modeling at equal compute ($10$K steps, $164$M tokens, single
seed). Test BPC; raw-text roundtrip; characters per token and NLL (nats) per
token on the test split; peak GPU memory during generation at batch $32$.
$^\dagger$Not exactly reversible on raw text; the Morfessor pipeline splits on
whitespace and was not roundtrip-tested.}
\label{tab:lm}
\end{table}

\begin{figure*}[t]
    \centering
    \includegraphics[height=3.4cm]{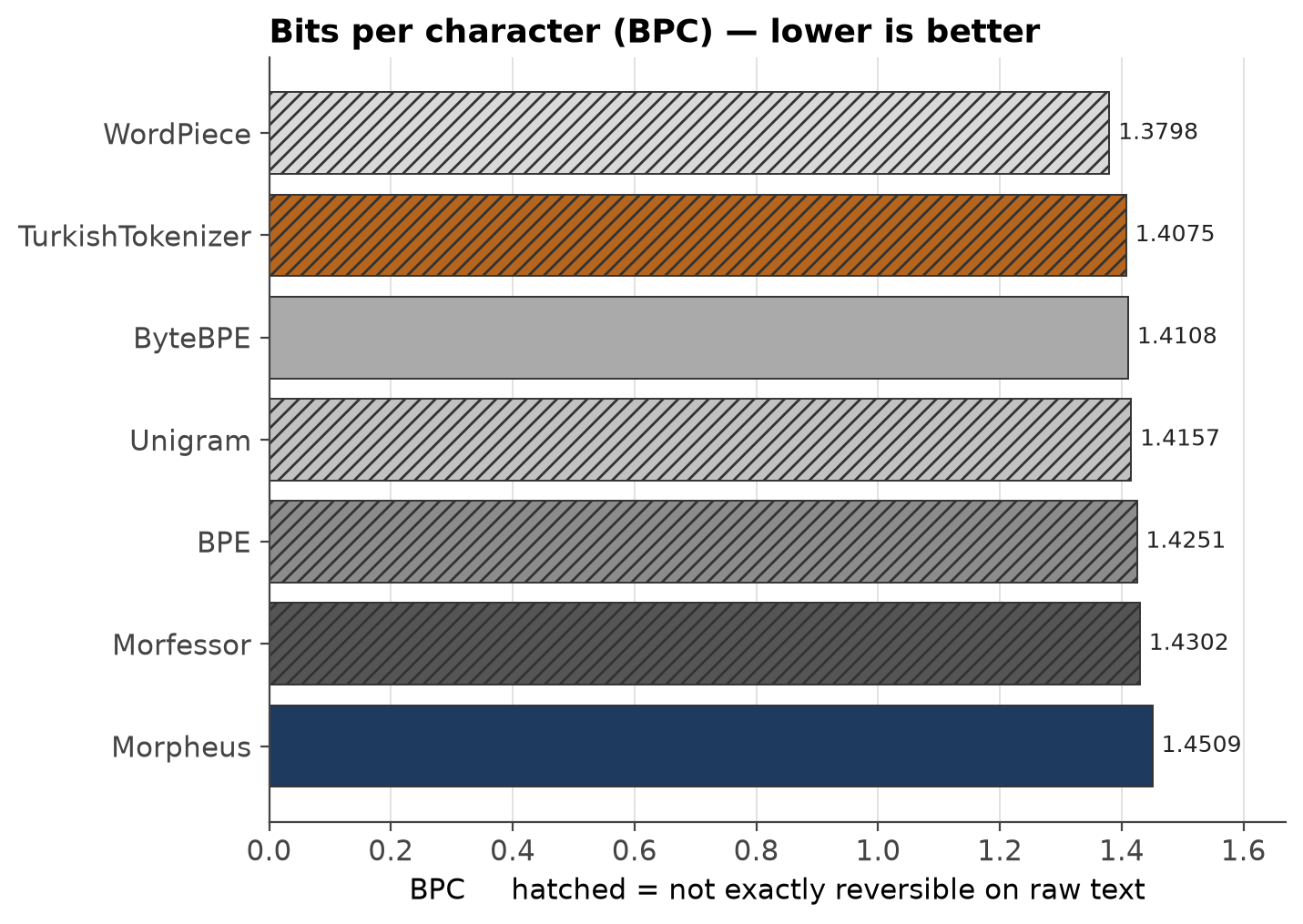}\hfill
    \includegraphics[height=3.4cm]{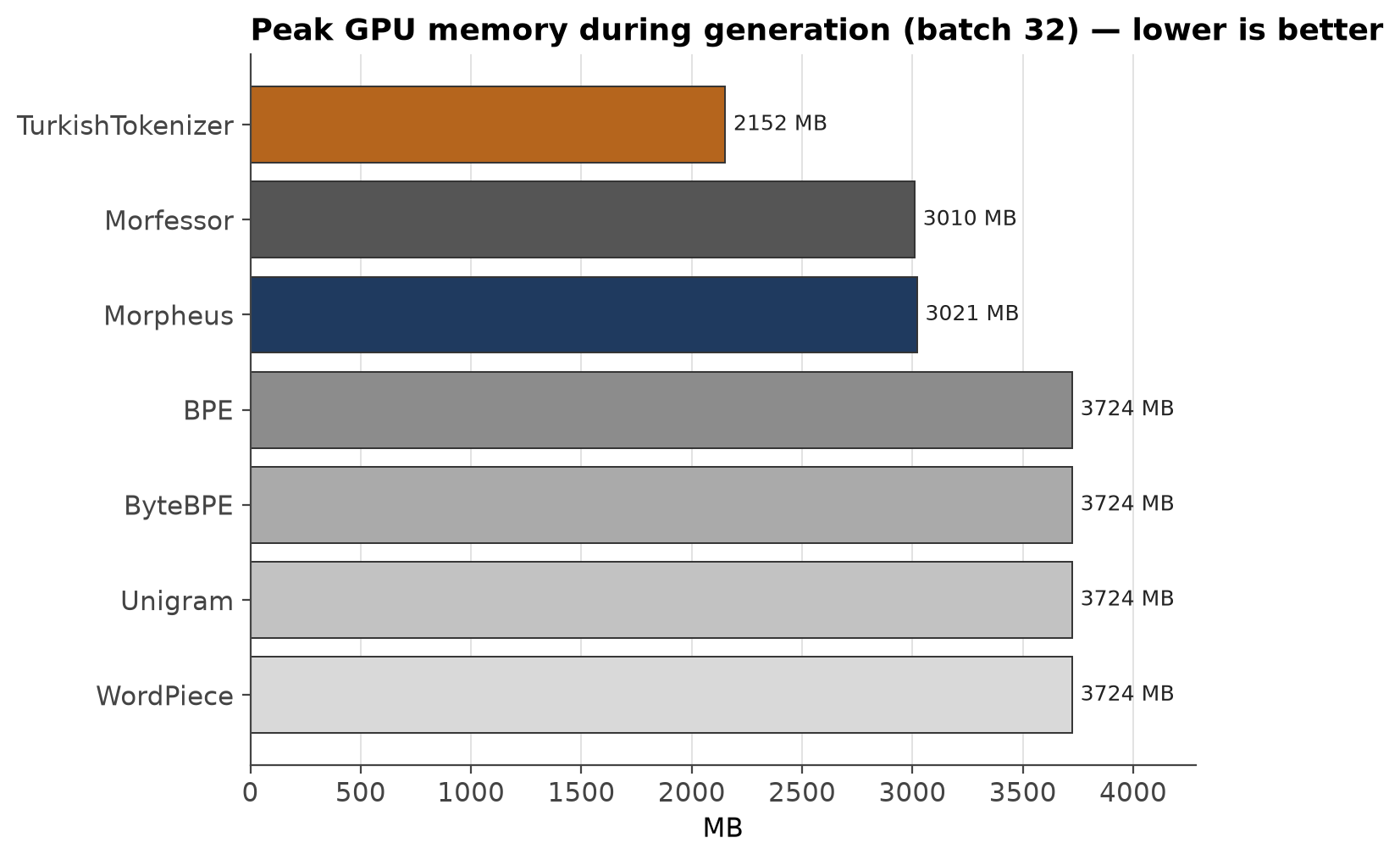}\hfill
    \includegraphics[height=3.4cm]{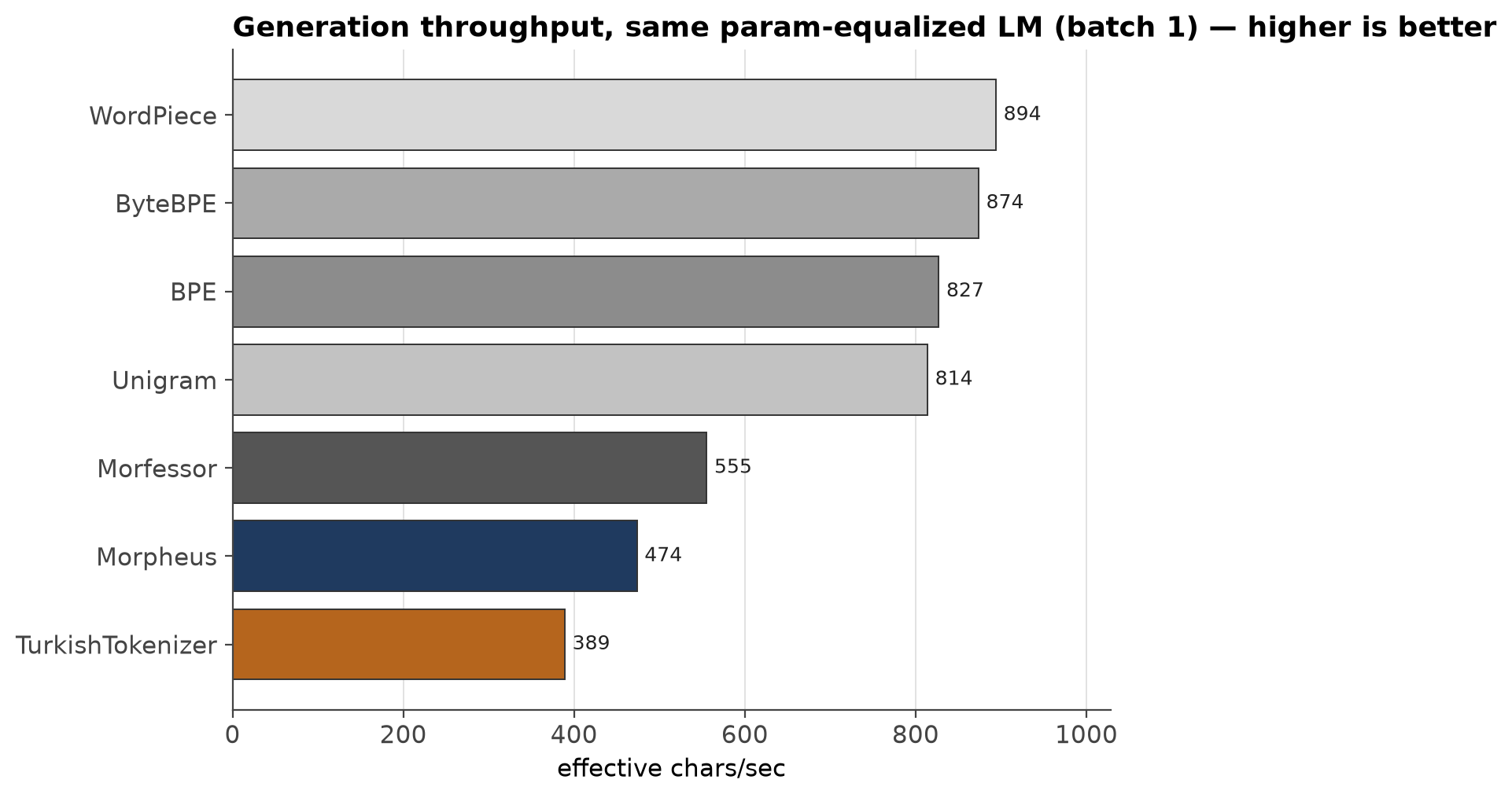}
    \caption{Language-modeling efficiency. Left: test BPC at equal compute
    (hatched: not exactly reversible on raw text). Middle: peak GPU memory during
    generation. Right: end-to-end generation throughput.}
    \label{fig:efficiency}
\end{figure*}

\subsection{Efficiency}
\label{sec:efficiency}
\begin{figure}[t]
    \centering
    \includegraphics[width=\linewidth]{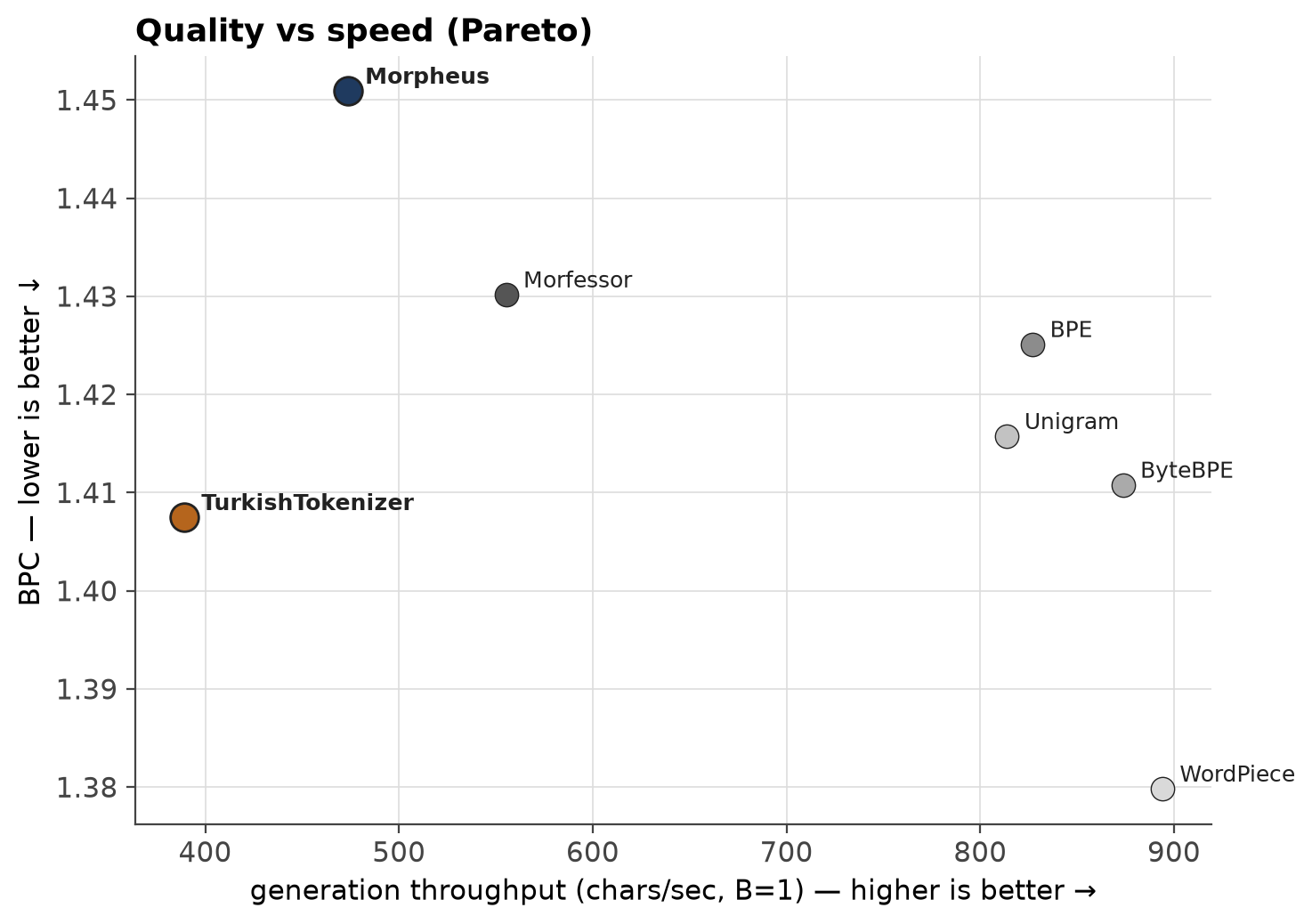}
    \caption{BPC versus generation throughput. On these two axes alone Morpheus is
    dominated by the subword tokenizers; its case rests on reversibility combined
    with morphological alignment and on its embeddings.}
    \label{fig:pareto}
\end{figure}

\paragraph{Generation.}
Because a Morpheus sequence is $30\%$ longer, generating the same text needs more
autoregressive steps: with the same $58$M model, Morpheus produces $474$
characters per second at batch $1$ against $874$ for byte-level BPE ($1.84\times$)
and $827$ for BPE, and $12{,}378$ against $21{,}787$ at batch $32$ ($1.76\times$;
Figures~\ref{fig:efficiency} and~\ref{fig:pareto}). Peak generation memory at
batch $32$ is $19\%$ lower for Morpheus ($3{,}021$ vs.\ $3{,}724$ MB). This saving
follows from vocabulary size---the embedding and output layers of a $50$K
vocabulary are smaller than those of a $64$K one, and the $32$K-vocabulary
TurkishTokenizer is lower still---so it would also be available to a $50$K
subword tokenizer and should not be credited to morphology.

\begin{figure*}[t]
    \centering
    \includegraphics[height=4cm]{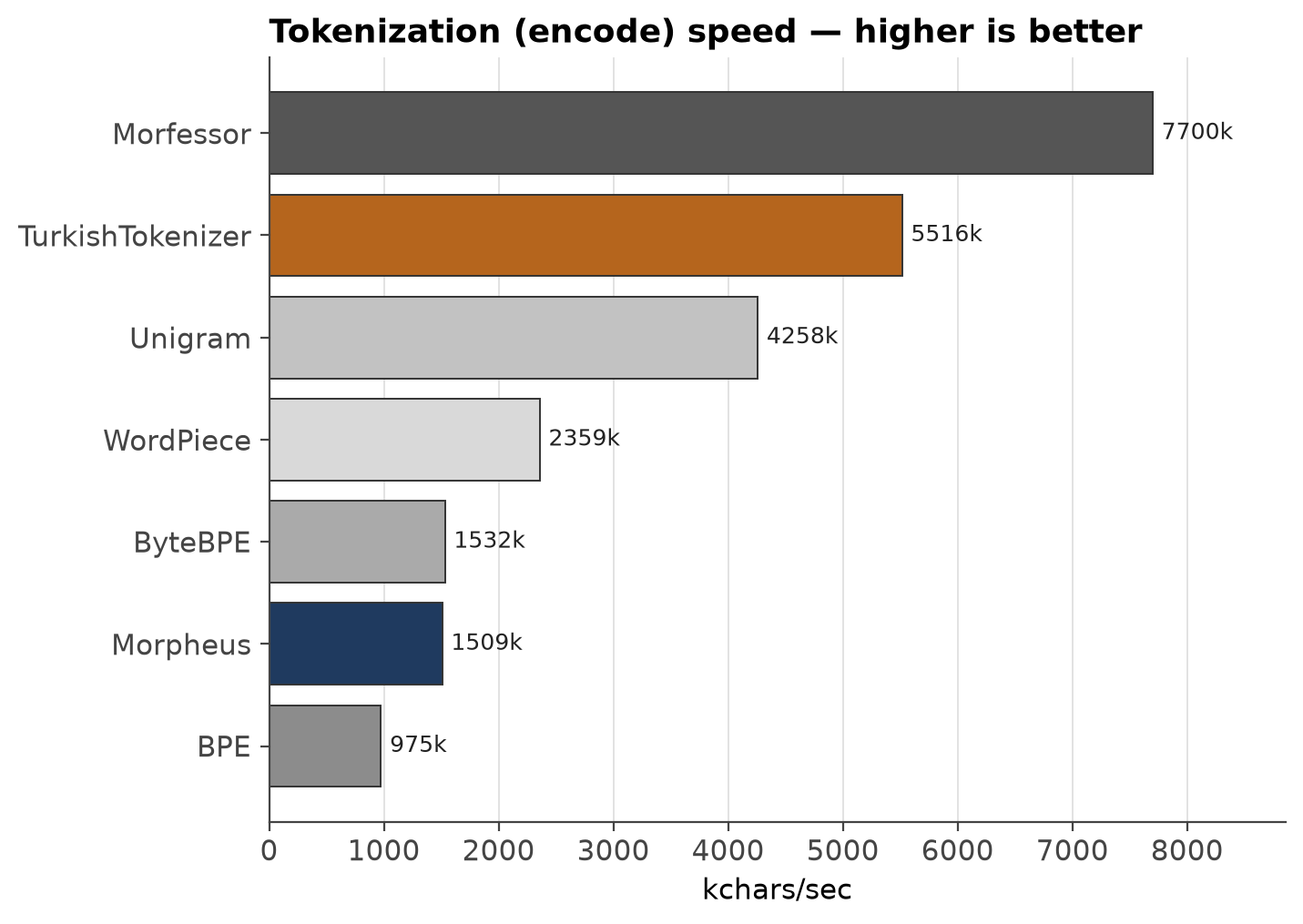}\hfill
    \includegraphics[height=4cm]{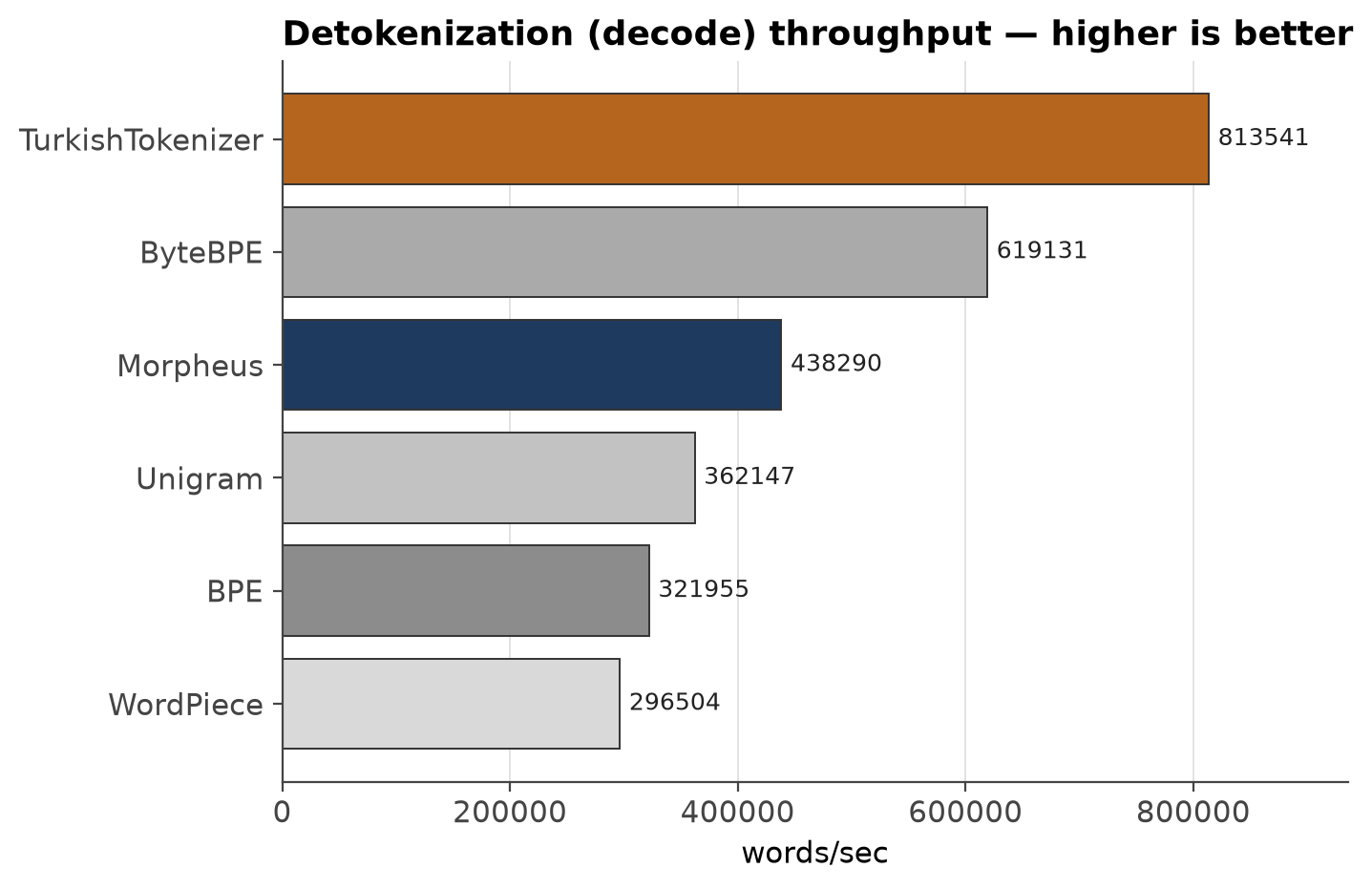}
    \caption{Tokenizer throughput, separate from end-to-end generation. Left:
    encoding speed (chars/s). Right: decoding speed (words/s).}
    \label{fig:encdec}
\end{figure*}

\paragraph{Tokenizer throughput.}
Morpheus's PyTorch encoder, including pre-tokenization and byte fallback, encodes
$1.51$M characters per second, on par with byte-level BPE ($1.53$M) and faster
than SentencePiece BPE ($0.98$M), but behind Unigram ($4.3$M), TurkishTokenizer
($5.5$M), and Morfessor ($7.7$M) (Figure~\ref{fig:encdec}). It decodes $0.44$M
words per second, faster than BPE, Unigram, and WordPiece ($0.30$--$0.36$M) and
behind byte-level BPE ($0.62$M) and TurkishTokenizer ($0.81$M). Both rates are
three orders of magnitude above generation speed, so tokenization is not the
bottleneck of a Morpheus-based model. The deployable tokenizer is larger: its
inference checkpoint and vocabulary occupy $29.6$ MB, against $1.3$--$4.6$ MB for
the subword tokenizers, and segmenting unseen words requires running the neural
model.

\subsection{Morpheus as a word embedder}
\begin{figure*}[t]
    \centering
    \includegraphics[height=3.6cm]{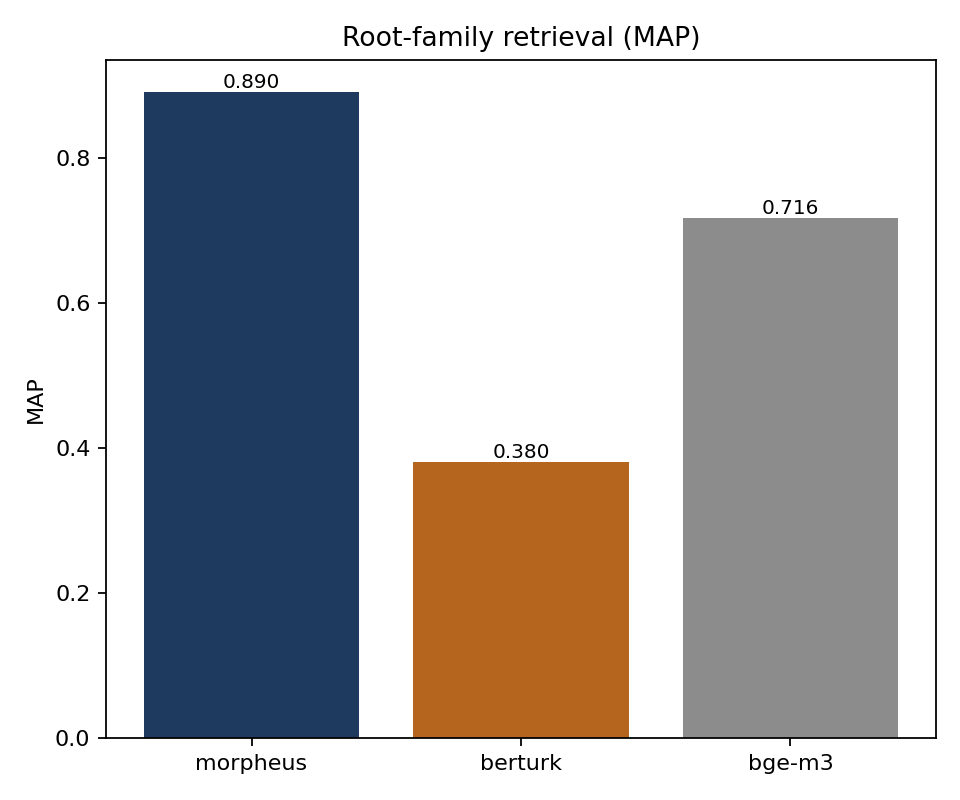}\hfill
    \includegraphics[height=3.6cm]{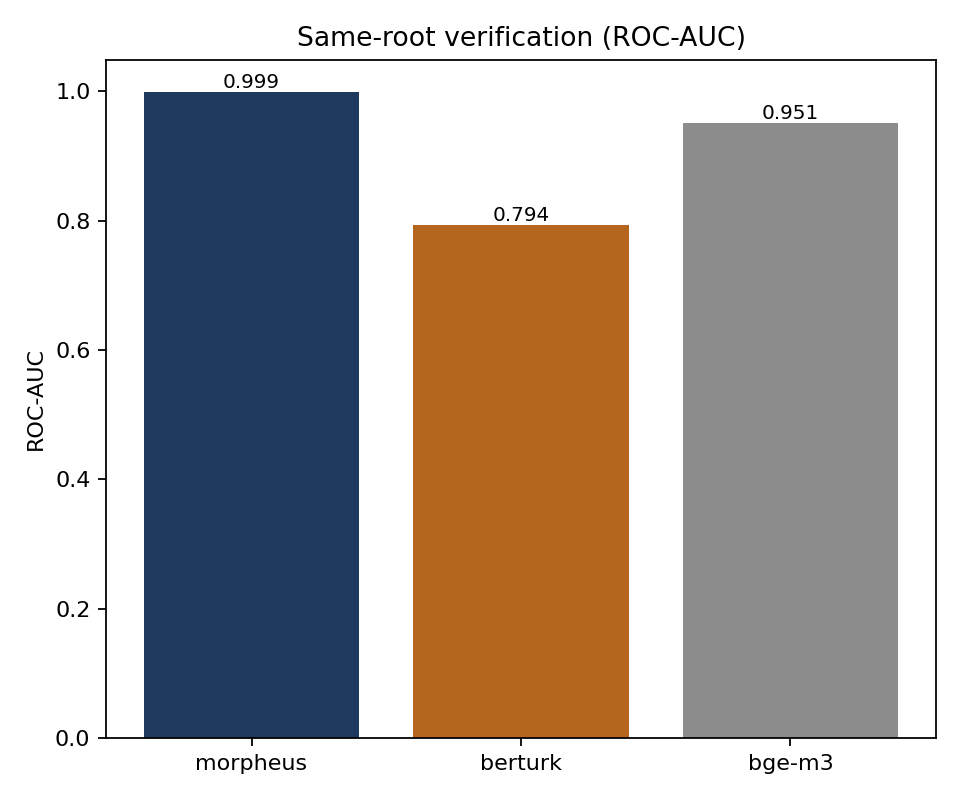}\hfill
    \includegraphics[height=3.6cm]{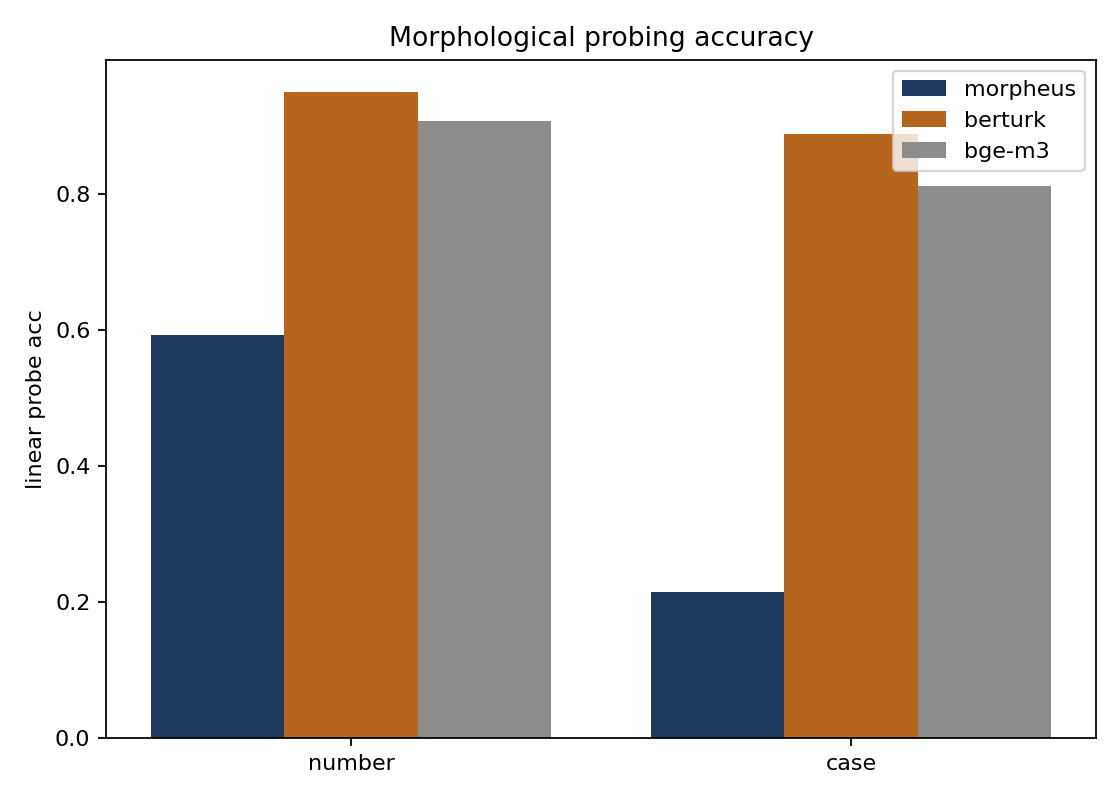}\hfill
    \includegraphics[height=3.6cm]{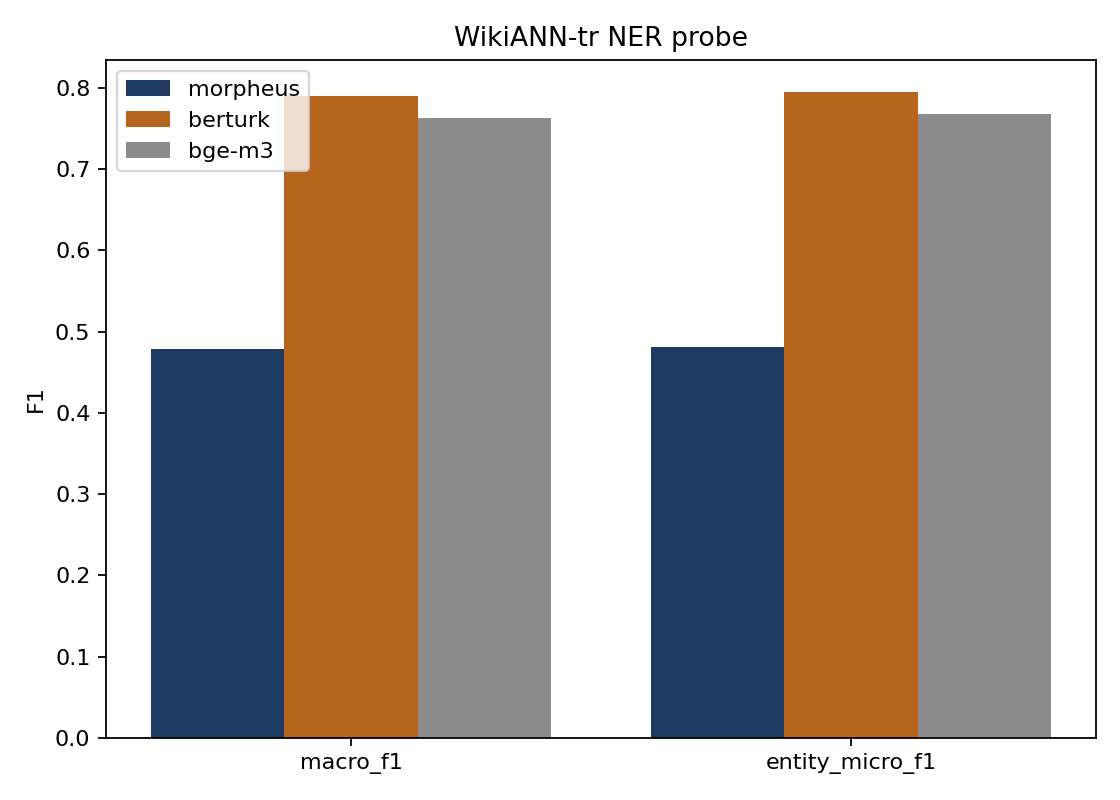}
    \caption{Embedding evaluation across encoders: root-family retrieval (MAP),
    same-root verification (ROC-AUC), morphological probing, and NER.}
    \label{fig:embed_bars}
\end{figure*}

Because Morpheus is neural, the same forward pass that tokenizes also emits a
$320$-dimensional word embedding. It is evaluated frozen against BERTurk and BGE-M3
(Table~\ref{tab:embed}, Figure~\ref{fig:embed_bars}); for both contextual models a
word vector is the mean of its subword states when the word is encoded alone.
Root families for retrieval and same-root verification are built from the gold
stems of UD\_Turkish-Kenet: $120$ families sampled at random from those with at
least eight distinct forms, $1{,}664$ words in total.

\begin{figure*}[t]
    \centering
    \includegraphics[height=5.3cm]{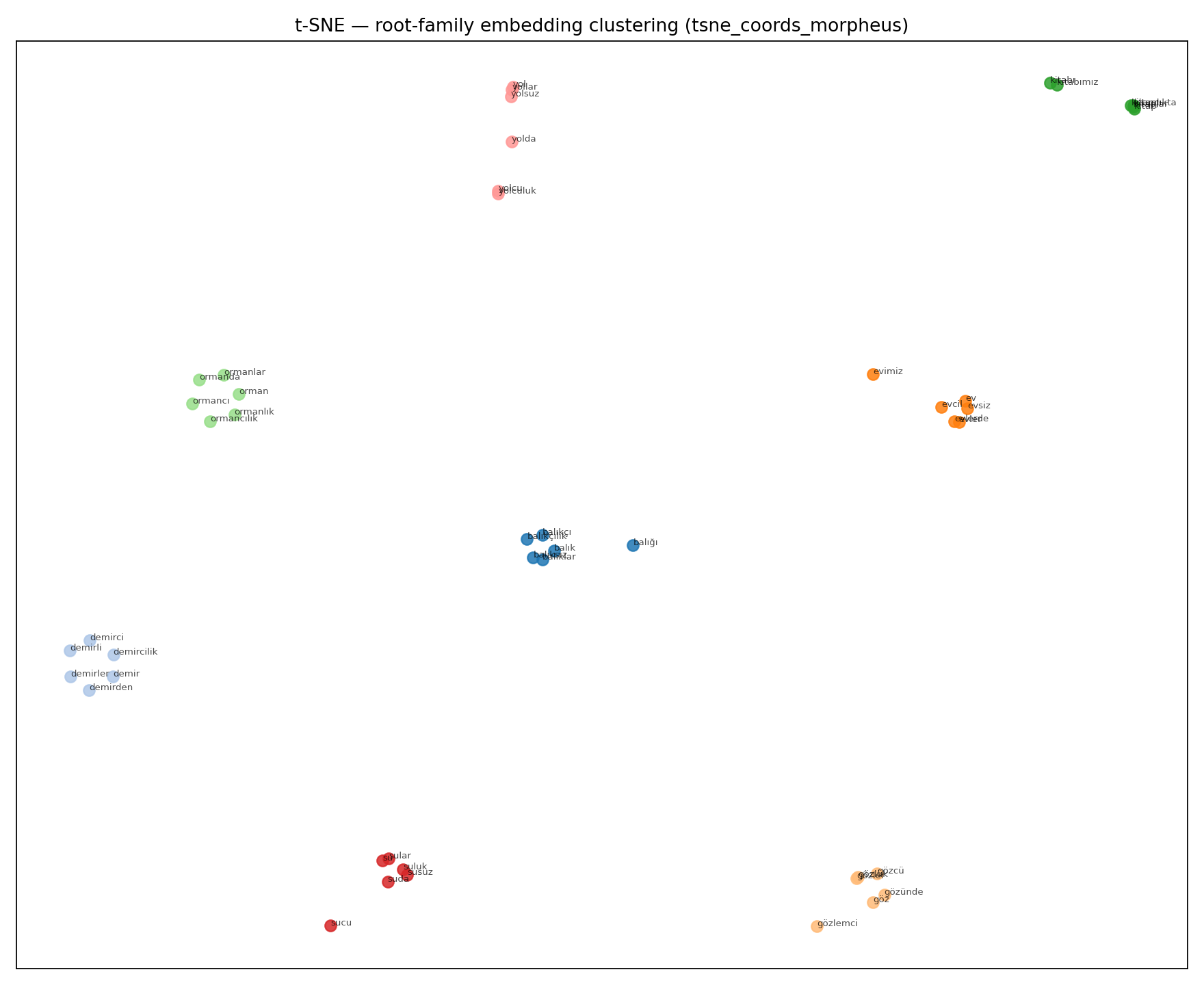}\hfill
    \includegraphics[height=5.3cm]{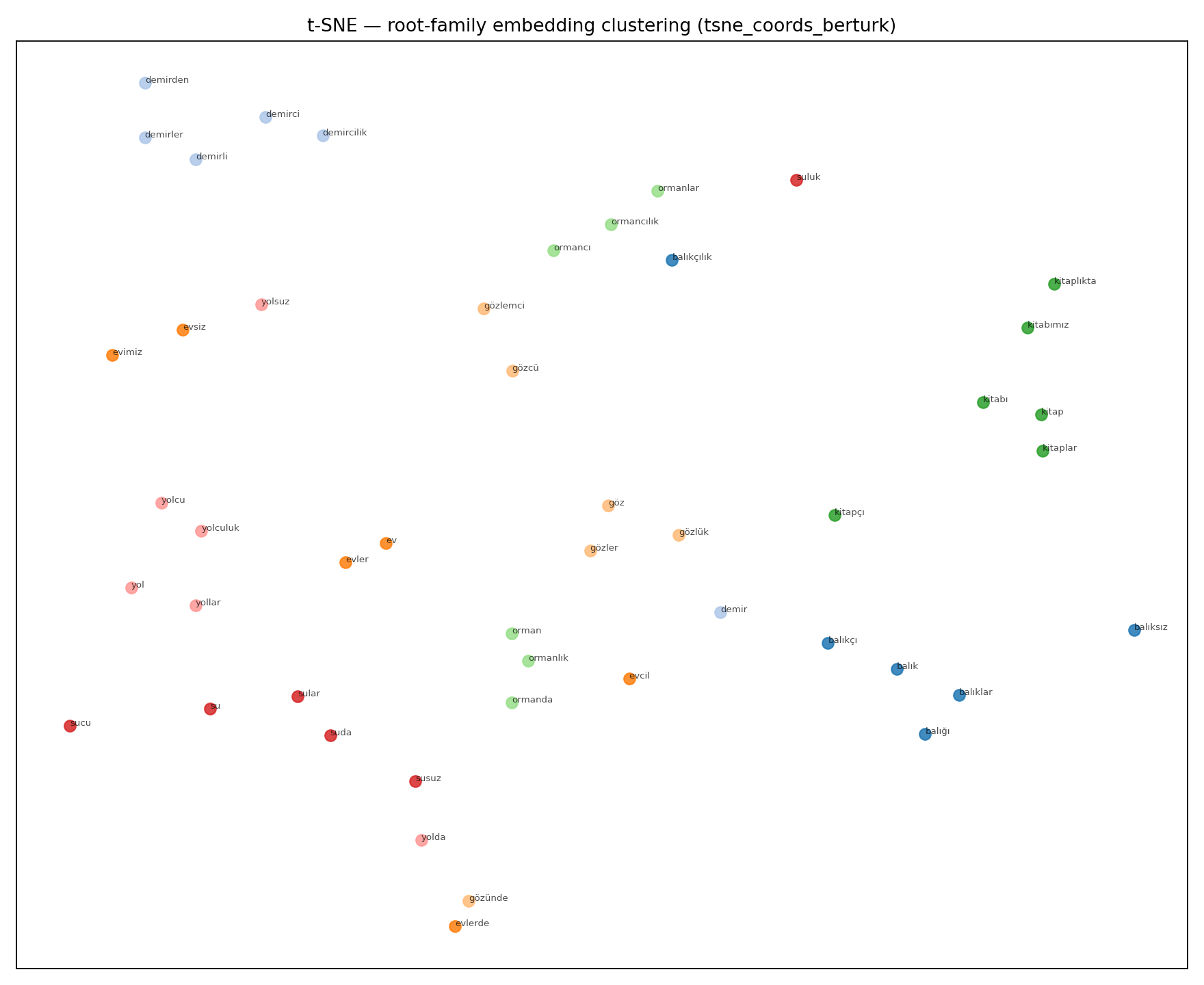}\hfill
    \includegraphics[height=5.3cm]{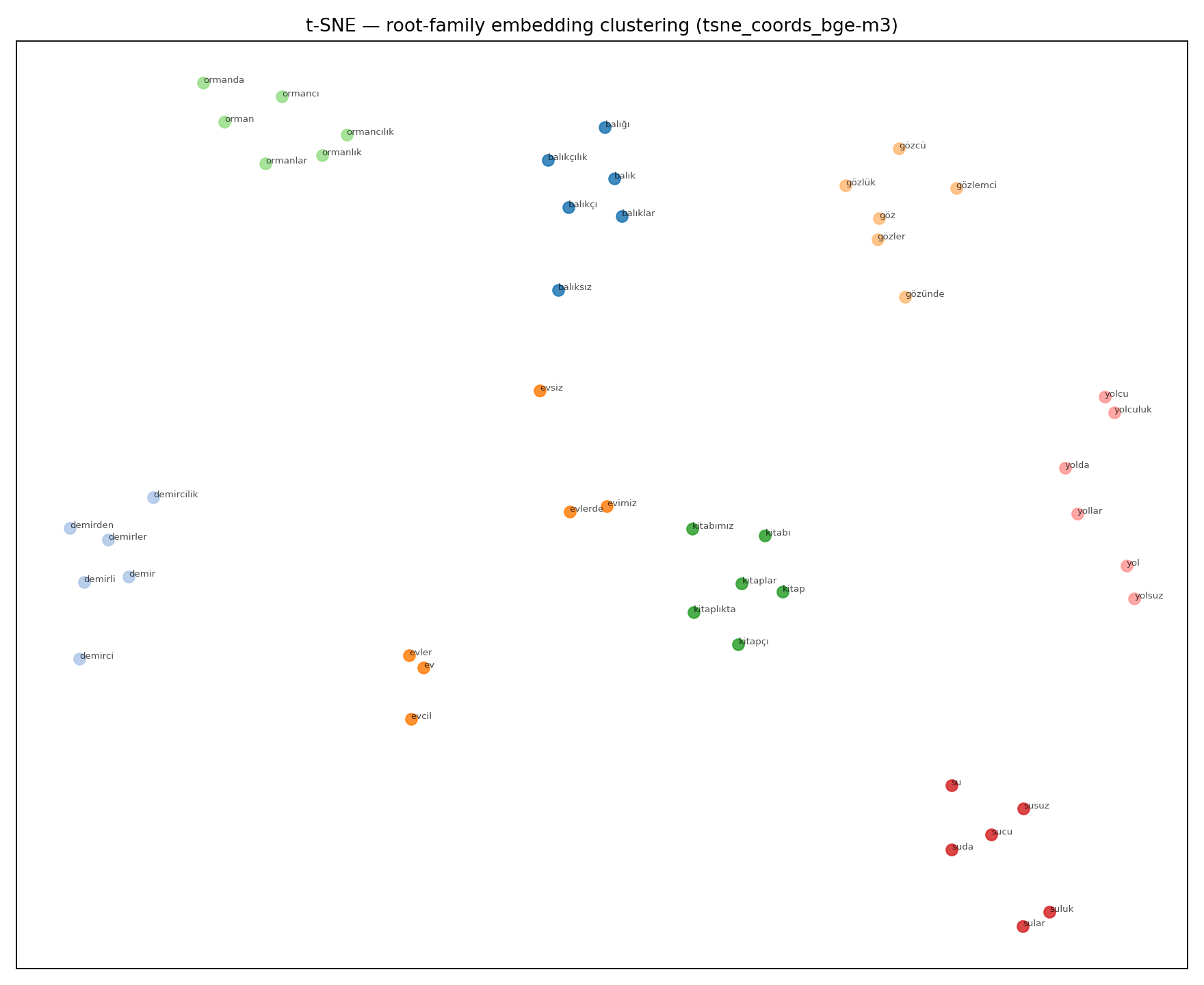}
    \caption{t-SNE of word embeddings for eight illustrative root families, for
    Morpheus (left), BERTurk (middle), and BGE-M3 (right). The quantitative
    evaluation uses the $120$ gold families of Table~\ref{tab:embed}.}
    \label{fig:embed}
\end{figure*}

\paragraph{Where Morpheus wins: lexical / root-level tasks.}
On retrieving other forms of the same root and on deciding whether two words
share a root, Morpheus leads decisively---root-family MAP $0.89$ (vs.\ $0.72$ for
BGE-M3 and $0.38$ for BERTurk) and same-root ROC-AUC $0.999$ (vs.\ $0.951$ and
$0.794$)---despite the smallest embedding ($320$ vs.\ $768$/$1024$ dimensions).
This is by design: the root-identity contrastive objective pulls all inflections
of a root toward a common point, and the t-SNE projections of eight illustrative
families (Figure~\ref{fig:embed}) show the resulting tight root clusters. The
comparison favors Morpheus in one respect that should be stated: its training
objective targets root identity directly, whereas BERTurk and BGE-M3 were not
trained for this task and are evaluated here out of context.

\paragraph{Where Morpheus loses: context- and inflection-dependent tasks.}
On morphological probing and NER the contextual encoders win clearly. A linear
probe on Morpheus vectors predicts number at $0.59$ (majority class $0.52$;
BERTurk $0.95$) and case at $0.22$ (majority class $0.20$; BERTurk $0.89$): the
embedding carries little number and almost no case information. On WikiANN NER
Morpheus reaches a macro-F1 of $0.48$ against $0.79$ for BERTurk. The same objective that sharpens root geometry
\emph{collapses} the inflectional contrasts a probe must read: by pulling
\textit{kitap}, \textit{kitaplar}, and \textit{kitab\i m\i z} together, it
discards the number and case signal that distinguishes them. The embedding is
also static, whereas NER is inherently contextual. Morpheus is therefore a
complementary, inexpensive \emph{lexical} encoder---suited to root matching,
deduplication, and keyword retrieval---rather than a replacement for a
contextual model.

\begin{table}[t]
\centering\small
\setlength{\tabcolsep}{4pt}
\begin{tabular}{lccc}
\toprule
 & Morpheus & BERTurk & BGE-M3 \\
 & (320) & (768) & (1024) \\
\midrule
Root-family MAP $\uparrow$ & \textbf{0.89} & 0.38 & 0.72 \\
Same-root ROC-AUC $\uparrow$ & \textbf{0.999} & 0.794 & 0.951 \\
Number probe $\uparrow$ ($0.52$) & 0.59 & \textbf{0.95} & 0.91 \\
Case probe $\uparrow$ ($0.20$) & 0.22 & \textbf{0.89} & 0.81 \\
NER macro-F1 $\uparrow$ & 0.48 & \textbf{0.79} & 0.76 \\
\bottomrule
\end{tabular}
\caption{Frozen word-embedding evaluation. Retrieval and verification use $120$
gold root families ($1{,}664$ words) from UD\_Turkish-Kenet; probing uses
SIGMORPHON number ($n=854$, $2$ classes) and case ($n=395$, $6$ classes) labels,
with majority-class accuracy in parentheses; NER uses WikiANN.}
\label{tab:embed}
\end{table}

\section{Discussion}

\subsection{Morpheus as a tokenizer: what it gains and what it costs}
\label{sec:assessment}
Table~\ref{tab:assessment} summarizes the tokenizer evidence against the properties a Turkish tokenizer
is usually judged on, using byte-level BPE---the only other exactly reversible
tokenizer in this comparison---as the reference.

\paragraph{Faithful decoding.}
Morpheus reconstructs every input at every tested level, including code,
emoji, irregular whitespace, and characters outside its vocabulary, and it never
emits an unknown token. On this property it is indistinguishable from byte-level
BPE. Tokenizers that normalize---even mildly, as NFKC does---fail on $1.3\%$ of
ordinary corpus lines and on half of the stress set.

\paragraph{Linguistic structure of the units.}
Among reversible tokenizers, Morpheus is the only one whose units align with
Turkish morphology: its MorphScore F1 is $0.17$ above byte-level BPE's, its unique
pure-token rate is $25$ points higher, and it segments rare roots exactly at
$1.6\times$ the rate of the best subword tokenizer. A recurring suffix such as the
locative \textit{-de/-da} or the plural \textit{-ler/-lar} is therefore presented
as the same unit across stems, rather than being absorbed into stem-specific
merges, and every segment is a surface string that can be inspected, counted, or
edited. This is what makes Morpheus useful for morphological analysis and
annotation, for lexical retrieval over morphological variants, and for any
application that needs to expose the morphemes of the text it processes.

\paragraph{Sequence length.}
The price of morpheme-level units is length: the same text is $30\%$ longer as a
Morpheus token sequence than as a byte-level BPE one ($1.90$ against $1.56$ tokens
per word on TR-MMLU). Any consumer of the token stream inherits this: a fixed
window of $512$ tokens spans $1{,}950$ characters for Morpheus and $2{,}546$ for
byte-level BPE, and an autoregressive model needs proportionally more steps to
produce the same text ($1.8\times$ in the generation benchmark).

\paragraph{The token stream as a modeling target.}
The controlled language-modeling probe of Section~\ref{sec:lm} asks how
predictable each token stream is for a small model under equal compute. Morpheus's
stream is $2.8\%$ harder to model than byte-level BPE's at $58$M parameters and
$10$K steps. Three candidate explanations are possible, none of which these data can
separate: (i) under an equal-token budget Morpheus sees $23\%$ less text; (ii) each
prediction conditions on less text, because the same context window holds fewer
characters; and (iii) a neural segmenter may cut the same root differently in
different words (Table~\ref{tab:decode}), which fragments statistics that a
deterministic merge table keeps together. The probe was run with a single seed at
a single small scale, so it characterizes the token stream rather than the
behavior of a full language model built on it.

\paragraph{Open vocabulary and deployment.}
Byte fallback gives Morpheus the same open-vocabulary guarantee as byte-level BPE,
while unseen Turkish words are segmented into morphemes rather than into
frequency-driven fragments. The price is operational: the tokenizer is a
$29.6$ MB neural model rather than a lookup table, and tokenizing new text
requires running it, at $1.5$M characters per second---on par with byte-level BPE
in these measurements.

\begin{table}[t]
\centering\small
\setlength{\tabcolsep}{3pt}
\begin{tabular}{lcc}
\toprule
Property & Morpheus & ByteBPE \\
\midrule
Exact decode (words / text / stress) & 100 / 100 / 14 & 100 / 100 / 14 \\
MorphScore macro-F1 & \textbf{0.61} & 0.44 \\
Rare-root exact segmentation & \textbf{70\%} & 10\% \\
TR-MMLU \%Pure (unique) & \textbf{57.1} & 32.1 \\
Tokens per word (TR-MMLU) & 1.90 & \textbf{1.56} \\
Characters per token & 3.81 & \textbf{4.97} \\
LM probe BPC (equal compute) & 1.451 & \textbf{1.411} \\
Encoding speed (chars/s) & 1.51M & 1.53M \\
Tokenizer artifact & 29.6 MB & \textbf{4.6 MB} \\
Word embedding by-product & \checkmark & -- \\
\bottomrule
\end{tabular}
\caption{Morpheus and byte-level BPE side by side, summarizing
Sections~\ref{sec:reversibility}--\ref{sec:efficiency}.}
\label{tab:assessment}
\end{table}

\paragraph{Assessment.}
As a tokenizer, Morpheus offers a combination that none of the tested
alternatives provides: exact reconstruction of raw text together with units that are
Turkish morphemes. It pays for this with longer token sequences, a heavier
artifact, and a token stream that a small language model finds slightly harder
to predict at equal compute. It is the better choice where surface fidelity and
morpheme-level units are both required---morphological segmentation and
annotation, linguistic and educational tools, retrieval and deduplication over
inflected forms, and preprocessing that must keep morphemes inspectable---while
byte-level BPE remains preferable where sequence length is the binding
constraint. Whether morpheme-aligned input improves a full-scale Turkish language
model trained on the same corpus is a separate question, which requires
equal-text and multi-seed comparisons, models of $\geq$1B parameters, and
downstream evaluation of the trained models; this is left to follow-up work.

\subsection{A root-centric embedding, by design}
The embedding results are a genuine finding rather than a shortfall. Morpheus
wins lexical retrieval and same-root verification but underperforms on
number/case probing and NER, and the cause is mechanistic: the contrastive
objective on root identity deliberately pulls all inflections of a root
together, sharpening root-level geometry while collapsing the inflectional
contrasts a linear probe would read, and the pooled static vector lacks the
sentence context NER needs. In a multi-vector retrieval system its embeddings are
a natural fit for the \emph{lexical} index, while a contextual model such as
BGE-M3 or BERTurk serves the dense semantic index.

\section{Limitations}
\label{sec:limits}

\paragraph{Scope of the language-modeling probe.}
All probe models have $58$M parameters, train for $10$K steps on $1$M lines, and
were run with a single seed. Differences of $0.005$--$0.014$ BPC among the
reversible subword baselines are within the range that seed variance could
produce, and the $0.040$ gap between Morpheus and byte-level BPE, while larger,
has not been confirmed across seeds or at equal text. The probe characterizes
the token streams; it does not measure how a full language model built on
Morpheus would behave.

\paragraph{Training data.}
The Morpheus checkpoint and the Morfessor model that supplied its training labels
were trained on an earlier snapshot of the corpus; the baselines, the Morpheus
vocabulary, the language models, and the evaluation splits use the rebuilt
corpus. Some held-out lines may therefore have been seen by Morpheus during
training. This can affect the evaluations derived from corpus words, but not the
external gold sets (UD\_Turkish-Kenet, SIGMORPHON, TR-MMLU, WikiANN), on which
the main morphological and embedding results rest.

\paragraph{Stem alternations and derivation.}
The surface-fidelity breakdown locates the boundary detector's main weaknesses:
stems that change under suffixation (consonant softening, vowel drop), loanword
harmony exceptions, where Morpheus finds the root but merges the following
suffixes, and derivational chains, where Morfessor is more accurate. Because
Morpheus never rewrites strings, these errors cost alignment but never
reversibility. The curated set is small ($50$ words), and the per-phenomenon
counts are indicative only.

\paragraph{Evaluation choices.}
The root-family retrieval task is aligned with Morpheus's contrastive training
objective, and the contextual baselines are evaluated on isolated words; the
retrieval results show what Morpheus's embedding is good at, not that it is a
better general-purpose encoder. The model's character inventory does not
distinguish \textit{q}, \textit{w}, \textit{x}, or circumflexed vowels, which
weakens segmentation of loanwords and foreign names (reversibility is
unaffected), and the Morfessor teacher was trained on whitespace tokens that
include attached punctuation.

\paragraph{Scope.}
The model and its supervision are Turkish-specific by design, and the gold sets
emphasize inflectional morphology, so derivational families and long, rare
agglutinative chains are comparatively under-probed.

\section{Conclusion}
Turkish agglutination exposes two gaps in the tokenizers that drive modern
language models: many cannot reproduce the text they encode, and those that can
are blind to the morphemes that carry meaning. This paper introduced
\textbf{Morpheus}, a neural morpheme-boundary model that closes both gaps at once
and, from the same forward pass, provides a structured word embedding. Its
central mechanism, a differentiable Poisson--binomial segmentation, lets
word-level objectives train the boundary detector end to end, recovers exact
segments at inference, and only \emph{groups} characters; with a lossless
pre-tokenizer and byte fallback this makes $\mathrm{decode}(\mathrm{encode}(x))=x$
hold for any text.

Under a comparison in which every baseline preserves case, Morpheus and
byte-level BPE are the only tokenizers that reconstruct every input, and Morpheus
is the only one of the two whose units are morphemes: MorphScore macro-F1 $0.61$
against $0.44$, a TR-MMLU pure-token rate of $57\%$ against $32\%$, and exact
segmentation of $70\%$ of rare roots. As an embedder it leads root-family
retrieval over $120$ gold families (MAP $0.89$). These benefits are bought with
$30\%$ longer sequences, $1.8\times$ slower generation, a heavier tokenizer
artifact, weaker handling of stem alternations, and, in the language-modeling
probe, a BPC $2.8\%$ above byte-level BPE's.

As a tokenizer, Morpheus is the only option in this comparison that is both
exactly reversible and morphologically aligned; as an embedder, it is a compact,
root-centric lexical encoder that complements rather than replaces contextual
models. Integrating Morpheus into a full Turkish language model trained on the
same corpus---with equal-text and multi-seed comparisons, larger models, and
downstream evaluation---is the subject of follow-up work.


\end{document}